%% file: main.tex
\documentclass[]{style}

\input{resources/packages}
\input{metadata}

\colorlet{OursRowColor}{WeChatGreen!12}
\newcommand{\oursrow}{\rowcolor{OursRowColor}}
\newcolumntype{C}{>{\centering\arraybackslash}X}
\newcommand{\ExpTableTextBold}[1]{{\fontsize{8.3}{9}\selectfont\bfseries\sffamily #1}}
\newcommand{\ExpTableSetup}{%
  \small
  \let\TableTextBold\ExpTableTextBold
  \setlength{\tabcolsep}{3pt}%
  \setlength{\aboverulesep}{0pt}%
  \setlength{\belowrulesep}{0pt}%
  \renewcommand{\arraystretch}{1.2}}

\usepackage{xspace}

\definecolor{codebg}{RGB}{248,248,248}
\definecolor{codecomment}{RGB}{100,100,100}
\lstdefinelanguage{json}{
  basicstyle=\ttfamily\small,
  numbers=none,
  breaklines=true,
  frame=none,
  backgroundcolor=\color{codebg},
  showstringspaces=false,
  literate= *{0}{{{\color{blue}0}}}{1} {1}{{{\color{blue}1}}}{1}
    {2}{{{\color{blue}2}}}{1} {3}{{{\color{blue}3}}}{1}
    {4}{{{\color{blue}4}}}{1} {5}{{{\color{blue}5}}}{1}
    {6}{{{\color{blue}6}}}{1} {7}{{{\color{blue}7}}}{1}
    {8}{{{\color{blue}8}}}{1} {9}{{{\color{blue}9}}}{1}
    {:}{{{\color{gray}:}}}{1} {,}{{{\color{gray},}}}{1}
}
\newtcblisting{jsonbox}{
  listing only,
  listing options={language=json,basicstyle=\ttfamily\scriptsize,
    breaklines=true,breakatwhitespace=false,columns=fullflexible,
    keepspaces=true,showstringspaces=false},
  colback=codebg,colframe=black!15,boxrule=0.5pt,arc=2pt,
  left=5pt,right=5pt,top=5pt,bottom=5pt
}

\AtBeginDocument{%
}

\let\reportsection\section
\renewcommand{\section}{\Needspace*{5\baselineskip}\reportsection}

\newcommand\blfootnote[1]{%
  \begingroup
  \renewcommand\thefootnote{}%
  \begin{NoHyper}%
    \footnotetext{#1}%
  \end{NoHyper}%
  \endgroup
}
\title{\modelname{}: Unified Visual Perception\\from Objects to Pixels}
\author{\ReportAuthorNames}
\affiliation{\ReportAffiliation}
\abstract{\ReportAbstract}
\DeclareRobustCommand{\frontlinkicon}[1]{%
  \raisebox{-0.3em}{\includegraphics[width=0.025\linewidth]{#1}}}
\DeclareRobustCommand{\frontresourcelink}[2]{%
  \ifstrempty{#1}{\todo{#2}}{\href{#1}{\texttt{#1}}}}
\checkdata[\frontlinkicon{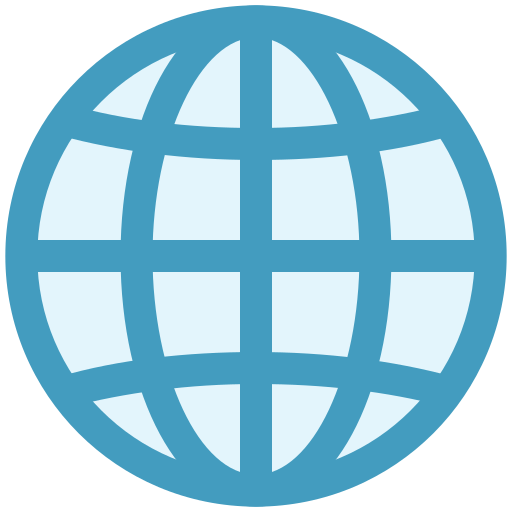}~~Project Page]{%
  \frontresourcelink{\ReportProjectPageURL}{Project page URL}}
\checkdata[\frontlinkicon{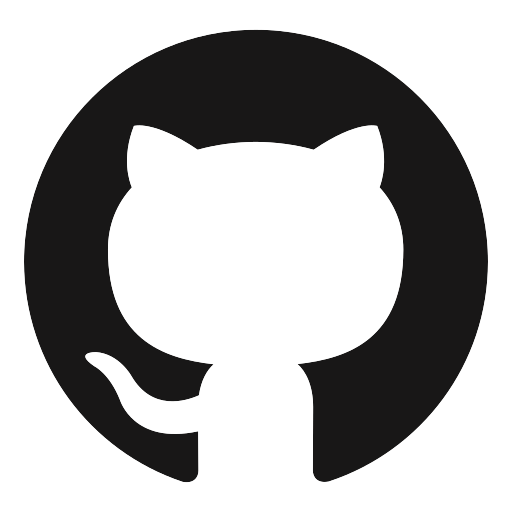}~~Code]{%
  \frontresourcelink{\ReportGitHubURL}{GitHub URL}}
\checkdata[\frontlinkicon{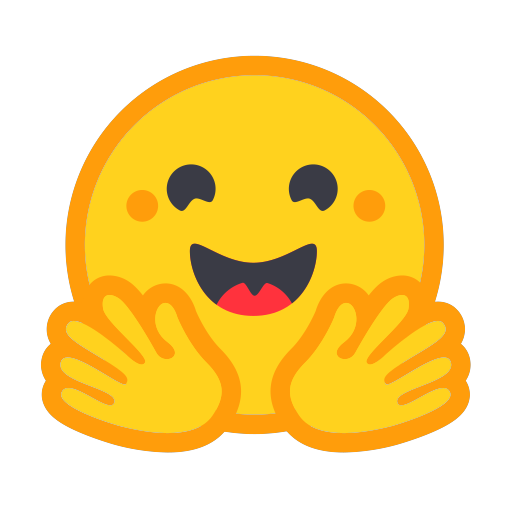}~~Weights]{%
  \frontresourcelink{\ReportHuggingFaceURL}{Hugging Face URL}}
\titleteaser{\input{sections/00_front_page}}

\begin{document}

\enlargethispage{1mm}
\blfootnote{\ReportContributionNote}
\maketitle

\input{sections/01_introduction}
\input{sections/02_data_engine}

\input{sections/03_method}
\input{sections/04_training}
\input{sections/05_experiments}
\input{sections/06_visualization}
\input{sections/07_related_work}
\input{sections/08_conclusion}

\clearpage
\bibliographystyle{plainnat}
\setlength{\bibhang}{0pt}
\setlength\bibindent{0pt}
\bibliography{references}

\clearpage
\input{sections/09_appendix}

\end{document}

%% file: resources/packages.tex
\usepackage{xargs}

\usepackage{multirow}

\usepackage{cleveref}

\usepackage{amsmath}
\usepackage{dsfont}
\usepackage{array}
\usepackage{longtable}
\usepackage{xurl}
\usepackage{needspace}
\usepackage{tikz}
\usetikzlibrary{arrows.meta,calc,positioning,shapes.geometric}
\usepackage{pgfplots}
\pgfplotsset{compat=1.18}
\usepgfplotslibrary{groupplots}

\usepackage{mathrsfs}
\usepackage{adjustbox}
\usepackage{multirow}
\usepackage{multicol}
\usepackage{tcolorbox}
\usepackage{changepage}
\usepackage{enumitem}
\usepackage{graphicx}
\usepackage{amssymb}
\usepackage{xcolor}
\usepackage{float}
\usepackage{multirow}
\usepackage{threeparttable}
\usepackage{graphicx}
\usepackage{wrapfig}
\usepackage[table]{xcolor}  
\usepackage{colortbl}       
\usepackage{tabularx} 
\usepackage{makecell} 
\usepackage[dvipsnames]{xcolor}

\newcolumntype{g}{>{\columncolor{gray!10}}c} 
\newcolumntype{A}{>{\raggedright\arraybackslash}p{4.5cm}} 
\newcolumntype{Y}{>{\raggedright\arraybackslash}X} 

\definecolor{catgray}{gray}{0.9}
\definecolor{skyblue}{rgb}{0.53,0.81,0.92} 

\colorlet{skyblue!30}{skyblue!30!white} 

\colorlet{themeboxborder}{WeChatDeepGreen}

\newtcolorbox{evolbox}[2][]{%
  enhanced,
  colframe=themeboxborder,
  colback=white,
  coltitle=white,
  rounded corners,
  boxrule=1pt,
  titlerule=0pt,
  toptitle=1mm,
  bottomtitle=1mm,
  fonttitle=\bfseries,
  width=#2\textwidth, 
  #1
}

\usepackage{url}

\PassOptionsToPackage{table,xcdraw}{xcolor}
\usepackage{titletoc}
\usepackage{placeins}
\usepackage{pifont}

\colorlet{RowTheme}{WeChatMint}
\definecolor{RowRed}{HTML}{F9EAEA}
\definecolor{Top1}{HTML}{50DB4B} 
\definecolor{Top2}{HTML}{A5FFA2} 
\definecolor{Top3}{HTML}{D9FFD9} 
\definecolor{Sub1}{HTML}{EAB8B8}
\definecolor{Sub2}{HTML}{E4E4E4}

\definecolor{gearred}{HTML}{D85140}
\definecolor{reprablue}{HTML}{5384ED}
\definecolor{steorange}{HTML}{EF8444}
\definecolor{softgreen}{HTML}{658E40}

\renewcommand{\emph}[1]{\textit{#1}}

\definecolor{codepink}{RGB}{220,20,120}
\definecolor{codegreen}{RGB}{0,150,0}
\definecolor{codegray}{RGB}{140,140,140}
\definecolor{codeorange}{RGB}{230,120,60}

\lstdefinestyle{pytorchstyle}{
    language=Python,
    basicstyle=\ttfamily\small,
    keywordstyle=\color{codepink}\bfseries,
    commentstyle=\color{codegray}\itshape,
    stringstyle=\color{codeorange},
    numberstyle=\tiny\color{codegray},
    numbers=none,
    showstringspaces=false,
    breaklines=true,
    frame=none,
    columns=fullflexible,
    keepspaces=true,
    xleftmargin=1.5em
}

%% file: metadata.tex
\providecommand{\todo}[1]{\textcolor{WeChatDanger}{\textbf{[TODO] #1}}}

\newcommand{\modelname}{WeVisOne-VL\xspace}

\newcommand{\ReportTitle}{\modelname}
\newcommand{\ReportOrganization}{WeChat Vision}
\newcommand{\ReportAuthorNames}{%
  Tianming Liang\textsuperscript{*},\quad Junyi Pan\textsuperscript{*},\quad Yuxi Li\textsuperscript{*},\quad Canmiao Fu,\quad
  Zhipeng Huang,\\[3pt]
  Chong Sun\textsuperscript{\ensuremath{\dagger}},\quad
  Chen Li\textsuperscript{\ensuremath{\ddagger}},\quad Jing Lyu%
}
\newcommand{\ReportAffiliation}{WeChat Vision, Tencent Inc.}
\newcommand{\ReportContributionNote}{%
  \textsuperscript{*}Equal contribution.\enspace
  \textsuperscript{\ensuremath{\dagger}}Project leader.\enspace
  \textsuperscript{\ensuremath{\ddagger}}Corresponding author.%
}

\newcommand{\ReportProjectPageURL}{https://tencent.github.io/WeVisOne-VL}
\newcommand{\ReportGitHubURL}{https://github.com/Tencent/WeVisOne-VL}
\newcommand{\ReportHuggingFaceURL}{https://huggingface.co/collections/tencent/wevisone-vl}

\newcommand{\ReportAbstract}{%
Existing vision-language models (VLMs) remain unreliable for fine-grained visual perception.
In this report, we present \textbf{\modelname{}} (WeChat Vision All-in-One), a powerful VLM that unifies object-level and pixel-level perception through structured next-token prediction. It supports multiple visual tasks including object detection, referring localization, counting, OCR, segmentation, and matting, enabling localization at any user-specific granularity, from points and boxes to masks and alpha mattes. 
Architecturally, we integrate a lightweight pixel decoder into the VLM via a special \texttt{<OBJ>} token to generate instance-level pixel outputs. We train the model with a three-stage curriculum SFT followed by mixed-task RL.
Across over 30 benchmarks, \modelname{} demonstrates leading performance on both object-level and pixel-level tasks, while maintaining robust rejection for object-absent queries.
By releasing \modelname{} to the open-source community, we hope to advance VLMs from broad visual understanding toward truly reliable perception systems for real-world deployment.
}

\hypersetup{
  pdftitle={\ReportTitle: Unified Visual Perception from Objects to Pixels},
  pdfauthor={Tianming Liang, Junyi Pan, Yuxi Li, Canmiao Fu, Zhipeng Huang, Chong Sun, Chen Li, Jing Lyu},
  pdfsubject={Unified object-level and pixel-level visual perception},
  pdfkeywords={vision-language model, perception, detection, referring, counting, OCR, segmentation, matting}
}

%% file: sections/00_front_page.tex
\noindent\begin{minipage}{\linewidth}
  \centering
  \includegraphics[width=\linewidth]{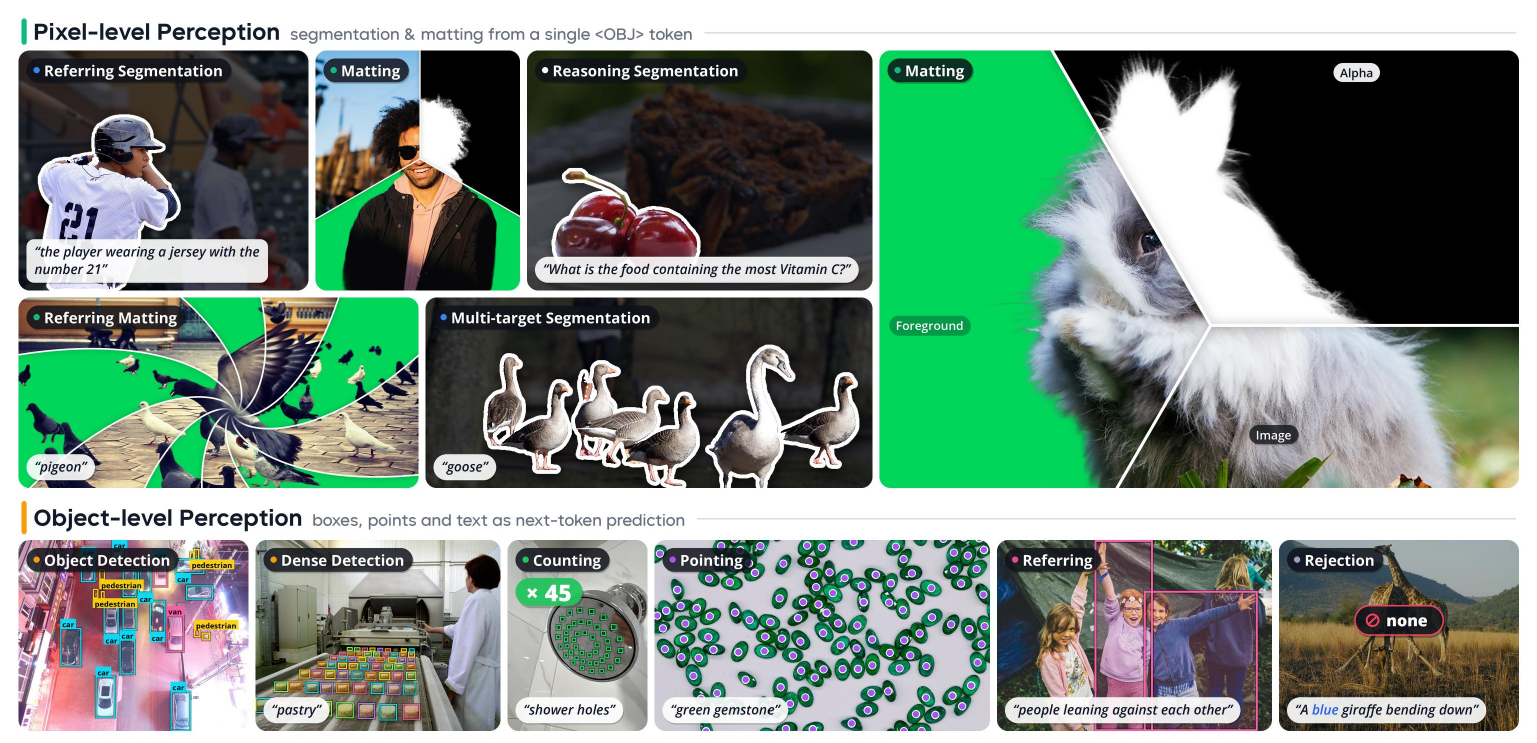}
  \captionsetup{hypcap=false,skip=2pt}
  \captionof{figure}{\small Overview of \modelname{}'s unified object- and
  pixel-level perception capabilities.}
\end{minipage}

%% file: sections/01_introduction.tex
\section{Introduction}
\label{sec:introduction}
Perceiving and understanding the visual world has long been a central pursuit of computer vision~\cite{krizhevsky2012imagenet,girshick2014rich,long2015fully,he2016deep,carion2020end,kirillov2023segment,bai2025qwen3,simeoni2025dinov3}. An intelligent visual system must be able to recognize what is present in an image, identify which object a user is referring to, and determine where these objects exist. Such capabilities establish the foundation for a wide range of real-world applications, such as content creation~\cite{liu2024referring}, industrial inspection~\cite{jeong2023winclip}, autonomous driving, and embodied intelligence~\cite{driess2023palme}.

Over the past decade, computer vision has made remarkable progress toward this goal. From conventional detection and segmentation models (e.g., YOLO~\cite{redmon2016you} and DETR~\cite{carion2020end}), to open-world perception models (e.g., Grounding-DINO~\cite{liu2024grounding} and SAM~\cite{kirillov2023segment}), and more recently to large Vision-Language Models (VLMs)~\cite{bai2025qwen3}, this field is being driven toward more unified, general, and powerful visual systems. This evolution has substantially broadened the concepts that a model can recognize and the instructions that it can follow. However, when both performance and efficiency are considered, current models are still far from satisfactory for real-world deployment. They often hallucinate non-existent objects, misinterpret user intent when encountering unfamiliar object descriptions, or overlook non-salient instances, leading to incomplete recall.

In this report, we introduce \textbf{\modelname{}} (\textbf{We}Chat \textbf{Vis}ion All-in-\textbf{One}), a frontier vision-language model that unifies object-level and pixel-level visual perception through a single natural-language interface. The model casts diverse visual tasks as structured next-token prediction and incorporates a lightweight pixel decoder for generating dense outputs.
Built on Qwen3-VL~\cite{bai2025qwen3}, we instantiate our models at two scales: 2B and 4B. We train the models on approximately 100M images with fine-grain annotations using a multi-stage curriculum. The key designs of \modelname are summarized as follows.

\textbf{Data Engine.} 
We develop a unified data engine that connects semantic discovery with instance-aligned supervision for object detection, referring localization, segmentation, and matting. To guide targeted acquisition and task-specific annotation, the engine organizes category concepts and compositional referring expressions in a shared knowledge graph. It then combines complementary localization experts to generate spatial candidates and consolidates their predictions through visual-language verification of query--instance correspondence. Building on these verified localizations, the engine generates binary masks to extend supervision to the pixel level while preserving target identity across tasks, with manual spot checks for additional quality control. For matting, it further leverages expert predictions and curated data to support a dedicated annotator that provides continuous alpha supervision. Finally, the engine uses reviewed model errors to guide subsequent acquisition, annotation correction, and hard-example sampling, enabling iterative refinement of the training data.

\textbf{Model Design.}
\modelname consists of a VLM backbone and a lightweight pixel decoder. Built on Qwen3-VL~\cite{bai2025qwen3}, \modelname casts a broad range of vision tasks as structured next-token prediction, generating object labels and spatial coordinates autoregressively. For compact and efficient output, we introduce a group of dedicated tokens (from \texttt{<0>} to \texttt{<1000>}) to encode normalized coordinates~\cite{jiang2026detect}. For dense prediction tasks, the model generates a special \texttt{<OBJ>} token after each predicted bounding box, whose hidden state serves as a latent representation of the corresponding instance. The pixel decoder combines this instance representation with multi-level features from the shared ViT to produce predictions over the full image canvas. In this way, \modelname supports both discrete and dense predictions without relying on a separate large segmentation model~\cite{lai2024lisa,yuan2026sa2va,liu2026unipixel,liang2026segresearch}.

\textbf{Training Pipeline.}
We train \modelname in a curriculum pipeline consisting of three-stage Supervised Fine-Tuning (SFT) followed by mixed-task Reinforcement Learning (RL). Stage 1 establishes task alignment, teaching the model to follow task instructions and produce well-formed structured outputs. Stage 2 introduces high-resolution and dense perception data to improve the recall of small and crowded objects. Stage 3 further incorporates pixel-level tasks, extending the model from discrete localization to dense prediction. For RL, we first select samples on which the SFT model is most uncertain, then apply Group Relative Policy Optimization (GRPO) over the mixture of tasks. This on-policy optimization effectively improves the localization accuracy, while suppressing repeated or hallucinated predictions.

Across multiple visual benchmarks, \modelname demonstrates highly competitive performance against both state-of-the-art general-purpose models and specialized vision systems. It exhibits consistently strong capabilities on both object-level and pixel-level tasks, including object detection, fine-grained referring, text recognition and language-guided pixel prediction. In addition, it also shows strong potential in challenging real-world scenarios, such as maintaining high recall in crowded scenes and correctly rejecting queries when the target is absent. By unifying these capabilities into a single model, we offer a powerful, efficient, and versatile vision-language foundation for the open-source community, accelerating the development of truly practical and reliable visual systems.

%% file: sections/02_data_engine.tex
\section{\modelname Data Engine}
\label{sec:data}

Data construction for unified visual perception must support both category-level recognition and context-dependent target selection, while preserving instance associations across language and spatial supervision. To this end, we build the \textbf{\modelname Data Engine} (WDE). WDE draws on GRIT~\cite{peng2023kosmos2}, CC3M~\cite{sharma2018conceptual}, LAION~\cite{laiondatasets}, SA-1B~\cite{kirillov2023segment}, and several task-specific datasets, and its annotation inventory comprises over 100 million images and approximately 3.6 billion bounding-box annotations. 

\begin{figure*}[t]
  \centering
  \includegraphics[width=\textwidth]{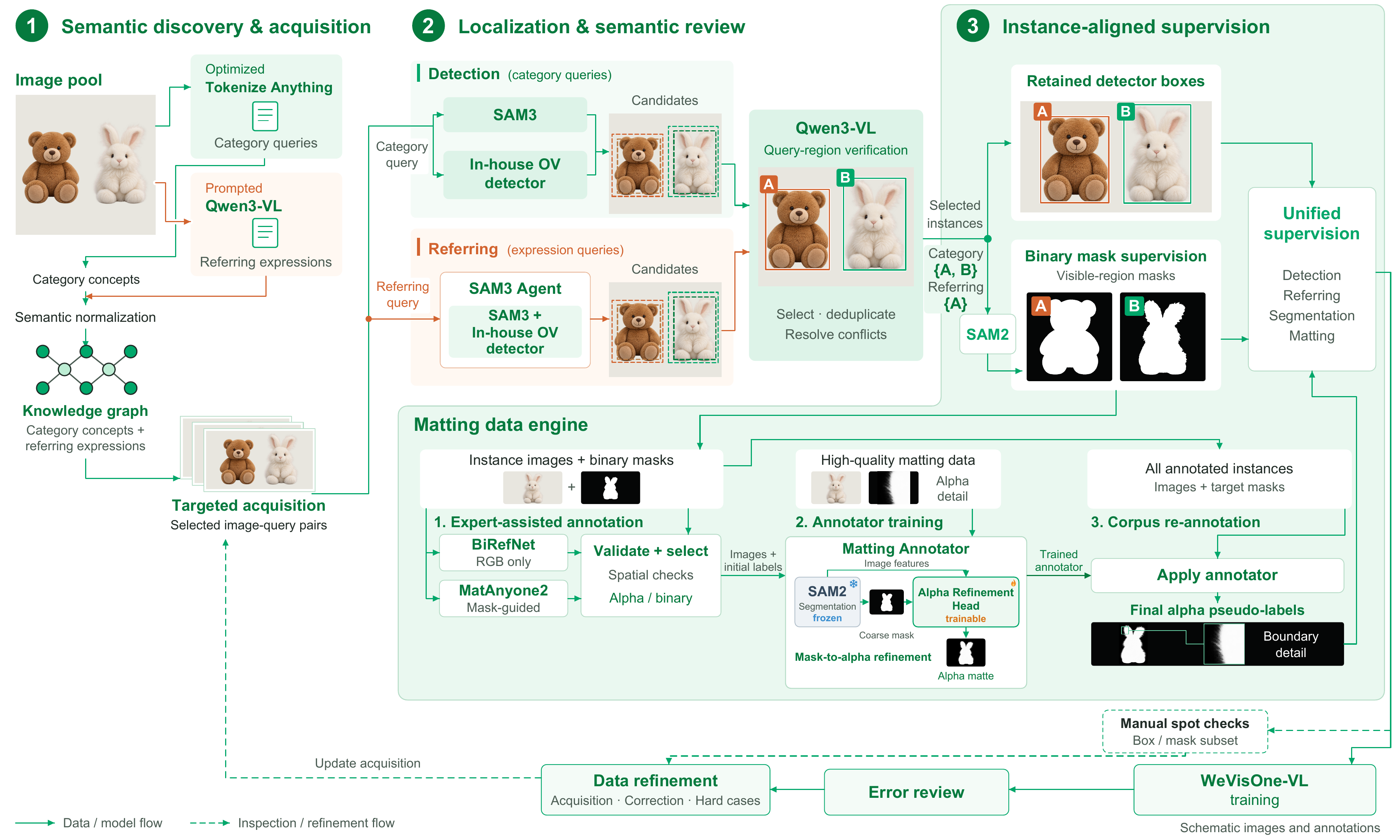}
  \caption{Overview of the \modelname Data Engine (WDE).}
  \label{fig:data_engine}
\end{figure*}

Figure~\ref{fig:data_engine} shows an overview of WDE. We first organize category concepts and referring expressions in a shared knowledge graph, which guides targeted image acquisition (Section~\ref{sec:data_discovery}). The acquired image--query pairs then go through task-specific localization and semantic verification, which establish the correspondence between each query and its target instances (Sections~\ref{sec:data_detection} and~\ref{sec:data_referring}). We keep these correspondences when generating binary masks (Section~\ref{sec:data_binary_masks}), and further extend the instance annotations to continuous alpha supervision with a dedicated matting pipeline (Section~\ref{sec:matting_data_engine}). Finally, model errors reviewed after each training round guide subsequent acquisition and annotation refinement (Section~\ref{sec:data_refinement}). Below, we describe each component in detail.

\subsection{Concept Discovery and Targeted Acquisition}
\label{sec:data_discovery}


Semantic discovery produces two complementary annotations: category-level concepts and compositional referring expressions. For detection, we use an optimized Tokenize Anything model~\cite{pan2024tokenize} to discover object categories. For referring tasks, we prompt Qwen3-VL~\cite{bai2025qwen3} to generate referring expressions that specify targets through attributes, states, actions, and relations. Category concepts support category-based localization, while referring expressions support selecting particular instances from their context. Both serve as annotation queries, and their correspondence to image regions is established in the subsequent localization stages.


We organize both category concepts and referring expressions in a shared knowledge graph. The graph is initialized from resources including Wikipedia\footnote{\url{https://www.wikipedia.org/}} and WordNet~\cite{miller1995wordnet}, and represents canonical category names, aliases, taxonomic relations, and whole--part relations. A candidate term is first linked to existing nodes through text retrieval and vision-language verification, and is considered for addition to the graph only when it is supported by stable visual evidence. During normalization, we merge equivalent category terms while keeping the descriptive constraints that determine the scope of referring expressions.


This semantic organization also supports targeted acquisition. Concept names and aliases serve as textual retrieval queries, verified instances provide visual retrieval cues, and historical errors indicate where additional coverage is needed. We set sampling priorities according to concept representation, image domain, target scale, and challenging conditions such as occlusion and dense object arrangements, and confusable categories and recurring failures in interpreting referring expressions further inform acquisition. In this way, annotation effort is directed toward both semantic gaps and visual conditions that remain difficult for the current model. The acquired image--query pairs are then sent to the detection or referring annotation pipeline.

\subsection{Detection Annotation}
\label{sec:data_detection}

Conventional object detectors excel at detecting densely distributed objects but struggle to recognize long-tail categories and open-world concepts. By contrast, VLMs offer strong recognition capabilities but often exhibit low recall in dense object scenarios. Therefore, for each object category, we first utilize detectors to generate exhaustive proposals, and then use a VLM for post verification. Specifically, we obtain proposals in parallel from SAM3~\cite{carion2026sam3} and an in-house open-vocabulary detector based on a DINO-style architecture~\cite{zhang2022dino}, which we refine iteratively. For each image--query pair, the predictions of both models are merged into a shared candidate pool. This design reduces our reliance on any single proposal source and uses the variation across detectors to pursue broader instance coverage.

We then use Qwen3-VL~\cite{bai2025qwen3} to verify whether each proposed region matches the category query given the image content. Based on its judgments, we select candidates, remove duplicates, and resolve conflicting predictions. For the retained instances, we keep the bounding boxes produced by the localization models. These boxes serve as detection supervision and also define the target instances used in the subsequent segmentation annotation.

\subsection{Referring Annotation}
\label{sec:data_referring}

Referring annotation requires the predicted regions to satisfy the compositional meaning of an expression, including the conditions that distinguish the intended targets within a scene. We build this stage on the SAM3 Agent framework~\cite{carion2026sam3} and add the in-house open-vocabulary detector as an additional localization expert. Within the same expression-conditioned workflow, SAM3 and the in-house detector both propose candidate regions, providing complementary localization evidence for the same expression.

Qwen3-VL then consolidates these proposals by checking whether each one matches the expression in its image context. The check covers the attributes, states, actions, and relations stated in the expression, and distinguishes the requested targets from objects that are mentioned only as contextual references. Whether an expression refers to a whole object or one of its parts is likewise determined by the expression itself. Conflicting proposals are resolved through the same semantic check. The accepted results bind each expression to an individual target or a target set, so that the annotation preserves the selection constraints of the query.

\subsection{Mask Annotation}
\label{sec:data_binary_masks}

For each localized instance, we use SAM2~\cite{ravi2025sam2} to predict a binary mask of its visible region. Each mask inherits the query of the box it is generated from. Each instance thus has both a box and a mask under the same query. With category queries, the boxes supervise detection and the masks supervise category-conditioned segmentation. With referring expressions, the boxes supervise referring localization and the masks supervise referring segmentation.

We manually spot-check a subset of these annotations, checking whether each mask and box are linked to the correct instance and whether the masks are accurate. Defects found in this check are corrected through the refinement process in Section~\ref{sec:data_refinement}. We also record the source of every annotation, so that affected labels can be traced and regenerated.

\subsection{Hard-Example Construction}
\label{sec:data_hard_examples}

Some distinctions may not be adequately represented by broad semantic coverage alone, and we construct hard examples to focus on them. Candidate queries are derived from confusable concepts in the knowledge graph, visually similar categories, controlled changes to descriptive conditions, and observed model failures. These sources provide plausible alternatives that can only be handled by precisely interpreting both the category and the contextual constraints of a query.

A query is negative only when no object in the image satisfies all of its conditions, and this cannot be decided from the query or the detector output alone. For example, changing ``red car'' to ``blue car'' does not mean the image contains no blue car, and a detector that returns nothing for a query may simply have missed the target. Therefore, we review each candidate negative against the whole image, checking the attributes, actions, and relations in the query. If a matching target turns out to exist, the query is sent back to positive annotation.

Whether an example is hard is determined by the behavior of the current model. A reviewed negative becomes a hard negative when the model nevertheless predicts a target region for it. Such examples carry an empty target set, so they teach the model to reject the query without introducing any box or pixel targets. We retain the erroneous predictions as evidence of the underlying confusion. Hard positives, in turn, capture missed targets, incorrect referent selection, and difficult localization under crowding or occlusion. We keep their reliable annotations, so that these examples strengthen the model's sensitivity to true targets while hard negatives teach it to reject. Sample difficulty is reassessed for each model version, and the sampling emphasis is adjusted toward persistent and newly emerging failure modes.

\subsection{Matting Data Engine}
\label{sec:matting_data_engine}

We build a dedicated matting pipeline within WDE to provide \modelname{} with alpha supervision across the diverse objects and scenes in our instance corpus. On the one hand, high-quality matting annotations are expensive to acquire, which limits their scale and coverage. Compositing foregrounds onto new backgrounds can enlarge the training set, but mismatched lighting and unnatural boundary transitions limit how well models trained on these images generalize to real scenes~\cite{li2022bridging,yang2026matanyone2}. On the other hand, our instance annotations already identify each target and its spatial extent, but their binary masks lack the continuous alpha values needed for detailed boundary supervision~\cite{yang2026matanyone2}. Extending them to matting therefore requires recovering fine details while preserving the target specified by each description, even when the reference mask has a coarse or inaccurate boundary. To this end, we combine expert-assisted annotation, dedicated annotator training, and corpus re-annotation, producing alpha labels that remain associated with the original instance descriptions and bounding boxes.

\textbf{Expert-assisted Annotation.}
For each instance with binary mask $S$, we obtain two alpha candidates on the same crop: the soft foreground mask $\alpha^{B}$ from BiRefNet~\cite{zheng2024birefnet} and the alpha matte $\alpha^{M}$ from MatAnyone2~\cite{yang2026matanyone2}. We initialize MatAnyone2 with $S$ after dilation and erosion, repeat the crop as a short frame sequence, and take the final alpha prediction after warmup. BiRefNet, in contrast, predicts foreground from the RGB crop without any instance prompt, so its output may correspond to a different object or cover only part of the intended target. We therefore use its binarized foreground $B=\mathbf{1}[\alpha^{B}\geq\tau_{\alpha}]$ only to check spatial consistency with $S$. Specifically, we set $\tau_{\alpha}=0.5$ and accept a BiRefNet candidate only if all of the following criteria hold: $|B|\geq8$, $\operatorname{IoU}(B,S)\geq0.5$, coverage $|B\cap S|/|S|\geq0.5$, foreground precision $|B\cap S|/|B|\geq0.8$, area ratio $|B|/|S|\in[0.5,2]$, bounding-box IoU of at least $0.7$, and a centroid offset no greater than $0.15$ times the box diagonal of $S$. For an accepted candidate, we further retain only the connected components of $\alpha^{B}\geq0.01$ that have at least $10\%$ of their area overlapping a slightly dilated $S$. The alpha values within these components are preserved without clipping to $S$, allowing fine structures to extend beyond the coarse boundary. If the BiRefNet candidate is rejected, or fewer than eight foreground pixels remain at threshold $\tau_{\alpha}$ after filtering, we fall back to $\alpha^{M}$ when $\operatorname{IoU}(\mathbf{1}[\alpha^{M}\geq\tau_{\alpha}],S)\geq0.9$; otherwise, we retain $S$. We record the selected source for each instance, together with any MatAnyone2 fallback to $S$ caused by small crops or inference failures, so that model predictions can be distinguished from binary fallbacks.

\textbf{Matting Annotator Training.}
We train a dedicated matting annotator by sampling instances with initial labels and high-quality matting data in equal proportions, combining broad instance coverage with detailed alpha supervision. Inspired by SAM2Matting~\cite{shen2026sam2matting}, we adopt a frozen segmentation backbone with a trainable matting head. Specifically, we augment a frozen SAM2.1-Base+ model~\cite{ravi2025sam2} with a new lightweight alpha refinement head. The head fuses multi-scale image features, prompt-mask guidance injected at each decoding stage, and local detail features from a shallow RGB stem, using top-down decoding to directly predict an alpha matte.
During training, prompt masks are obtained by binarizing the target alpha, taking the instance segmentation mask, randomly perturbing these masks, or segmenting the target with box-prompted SAM2. The objective combines multi-scale $L_1$ and Laplacian alpha losses with foreground/background supervision and a prompt-consistency term applied only to unperturbed object-level prompts. Additional transition-region losses are applied to samples with continuous alpha labels, excluding small targets.

\textbf{Corpus Re-annotation.}
Finally, we apply the trained annotator to all annotated instances in the training corpus, using the existing instance annotations to specify each target. Its predictions replace the initial labels and serve as the final alpha pseudo-labels for the matting training described in Section~\ref{sec:training_pipeline}. Since each matte remains linked to its original instance description and bounding box, language grounding and matting are supervised on the same target.

\subsection{Training Feedback and Data Refinement}
\label{sec:data_refinement}

After each training round, we analyze model errors to guide both data acquisition and annotation refinement. We organize the errors by concept, image domain, target scale, occlusion, task, and failure type, and apply clustering and deduplication to select representative cases for review. This concentrates annotation effort on recurring failures and coverage gaps.

The review distinguishes errors in model prediction from errors in supervision. If the annotation is reliable, we keep the example unchanged; it may receive greater sampling emphasis or serve as a retrieval seed for related images. If the review identifies an annotation defect, we correct or regenerate the affected supervision before including it in a subsequent data version. In other words, model disagreement only triggers an investigation, and annotations are changed only after review.

Different failure modes lead to different updates. Missed targets guide additional acquisition and hard-positive sampling. Referent confusion motivates instance contrasts and more descriptive queries, while false positive responses contribute to hard-negative construction. Defects in boxes, masks, or cross-task instance associations are returned to the corresponding annotation stage for correction.

Each data release records the annotation provenance, the model version that exposed each error, and the associated review and revision history. Unresolved cases are excluded from training. Original images, their identified near duplicates, and the derived task records stay within the same data partition across updates. Reviewed feedback data can enter subsequent training versions, whereas a fixed monitoring set is reserved for internal regression and excluded from both training and threshold calibration. Formal benchmarks are kept isolated from training and from feedback-driven data construction, so that evaluation remains independent as WDE evolves.

%% file: sections/03_method.tex
\section{Model Design}
\label{sec:model_design}
\modelname{} is built upon Qwen3-VL~\cite{bai2025qwen3} with two key extensions. (1) \textbf{Task formulation}: it introduces dedicated tokens for normalized coordinates and task-specific tags, yielding a compact and efficient output format. (2) \textbf{Model architecture}: it incorporates a lightweight decoder that transforms discrete token representations into dense predictions for segmentation and matting. Below, we provide the details of these two designs.

\subsection{Task Formulation}
\label{sec:task_formulation}
\begin{figure*}[!ht]
  \centering
  \caption{Representative input--output examples for the unified task formats used by \modelname{}.}
  \label{fig:task_formulation}
  \includegraphics[width=\textwidth]{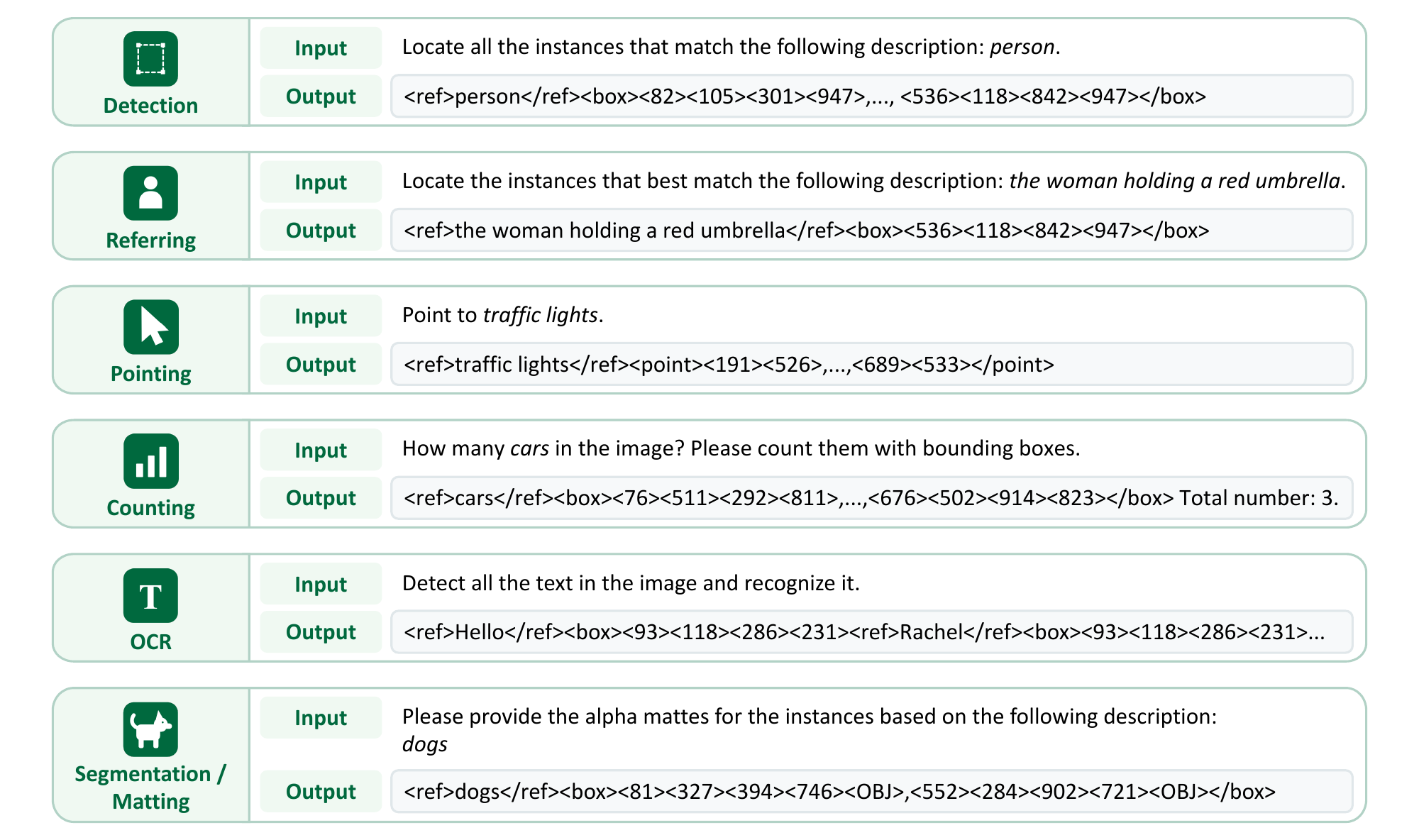}
\end{figure*}

We formulate all visual tasks as structured next-token prediction. Inspired by Rex-Omni~\cite{jiang2026detect} and LocateAnything~\cite{wang2026locateanything}, we introduce a group of dedicated tokens from \texttt{<0>} to \texttt{<1000>} to represent normalized coordinates, enabling higher token efficiency. As shown in Figure~\ref{fig:task_formulation}, \modelname{} localizes queried targets by sequentially generating object references and spatial coordinates. For tasks requiring bounding boxes, we encode the spatial coordinates as $[x_0,y_0,x_1,y_1]$. For tasks requiring point outputs, we use $[x_0,y_0]$ as the coordinates. If a queried target is absent from the image, its coordinate field is filled by \texttt{None}. For pixel-level tasks such as segmentation and matting, \modelname{} additionally generates an \texttt{<OBJ>} token immediately after each box-coordinate quartet, inside the same \texttt{<box>} field, to provide the target representation for pixel decoding.

\subsection{Model Architecture}
\label{sec:model_architecture}

Given an image and an instruction, \modelname{} first encodes the image with the Qwen3-VL vision encoder and autoregressively generates the structured response described above. For pixel-level tasks, it invokes a lightweight decoder for each generated \texttt{<OBJ>} token. The decoder takes the token's hidden state, its predicted box, and multi-level features from the Qwen3-VL vision encoder. The decoder draws image context from the shared visual features and spatial guidance from the predicted boxes, eliminating the need for a separate large segmentation model such as SAM~\cite{kirillov2023segment}.

\begin{figure*}[t]
  \centering
  \includegraphics[width=\textwidth]{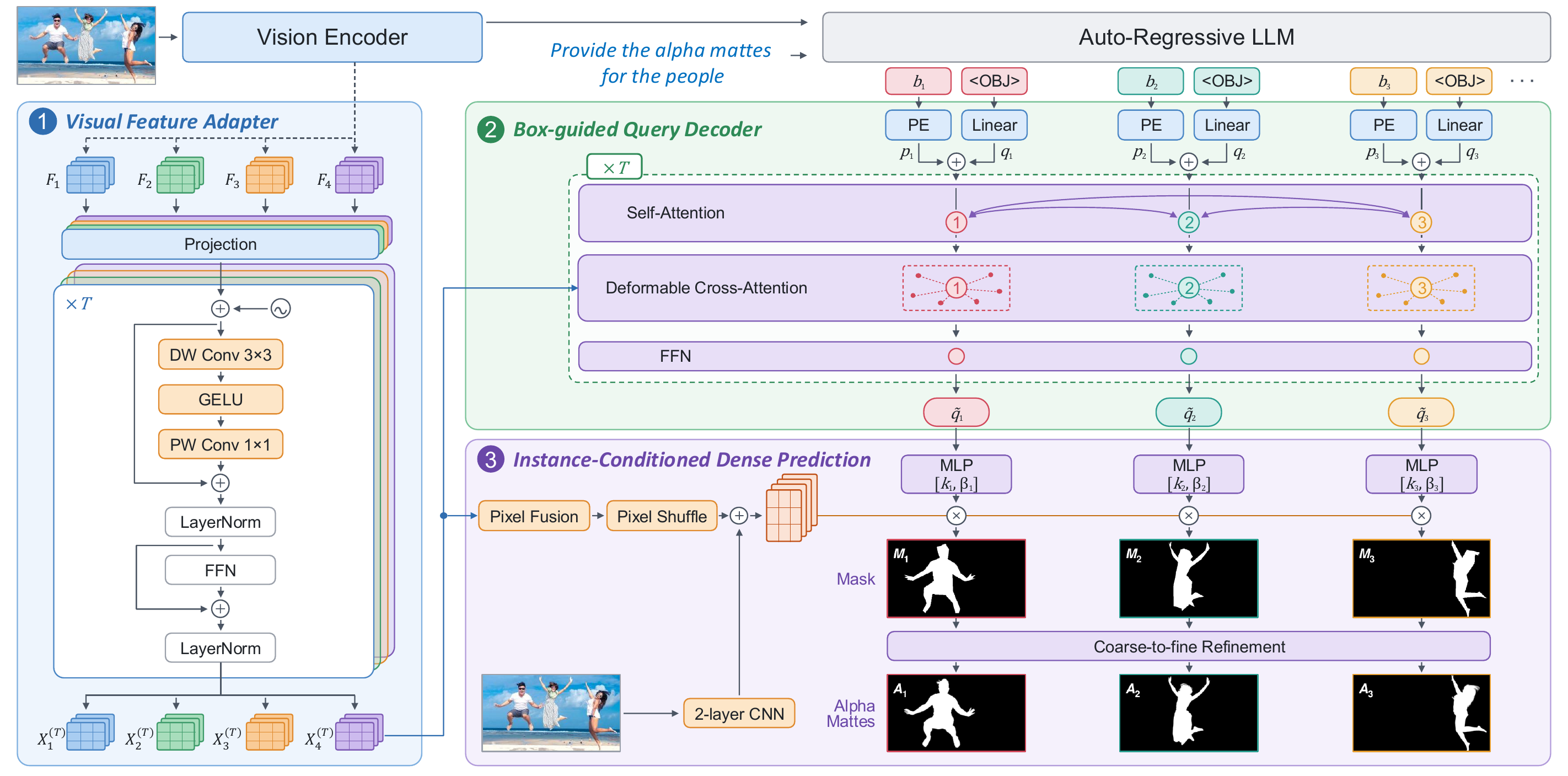}
  \caption{Overview of the lightweight pixel decoder. Qwen3-VL provides multi-level ViT features, the hidden state of each generated \texttt{<OBJ>} token, and its predicted box. The decoder builds shared visual features, updates a box-guided target query, and converts the two representations into a full-resolution mask or alpha matte.}
  \label{fig:pixel_decoder}
\end{figure*}
\FloatBarrier

The decoder consists of three components: a visual feature adapter that organizes intermediate ViT features into a shared multi-level memory, a deformable query decoder that combines each \texttt{<OBJ>} representation with its box prior, and an instance-conditioned dense prediction stage that converts the updated query and visual features into a full-image prediction. Segmentation and matting share these components, while matting uses an additional lightweight refinement module for fine boundaries and fractional opacity. The entire decoder, including the refinement module, contains about 9.6M parameters.

\subsubsection{Visual Feature Adapter} To preserve fine-grained visual information for dense prediction, we extract $L$ pre-merging features $\{F_l\in\mathbb{R}^{C\times H_p\times W_p}\}_{l=1}^{L}$ from the intermediate ViT layers, serving as multi-level visual features. For each level $l$, the features are mapped to a $d$-dimensional decoder space and combined with positional and level information:
\begin{equation}
  X_l^{(0)}=W_lF_l,
  \qquad
  E_l=P_{\mathrm{2D}}+e_l,
\end{equation}
where $W_l\in\mathbb{R}^{d\times C}$ is a learnable projection, $P_{\mathrm{2D}}\in\mathbb{R}^{d\times H_p\times W_p}$ is the 2D positional encoding, and $e_l\in\mathbb{R}^{d}$ is a learnable level embedding. The visual feature adapter is composed of $T$ blocks. Given the input $X_l^{(t)}$, the $t$-th block computes as follows:
\begin{equation}
\begin{gathered}
  \bar X_l^{(t)}
  =\operatorname{LayerNorm}\!\left(
    X_l^{(t)}+\operatorname{PWConv}_{t}\!\left(
    \operatorname{GELU}\!\left(
    \operatorname{DWConv}_{t}(X_l^{(t)}+E_l)\right)\right)\right), \\
  X_l^{(t+1)}
  =\operatorname{LayerNorm}\!\left(
    \bar X_l^{(t)}+\operatorname{FFN}_{t}(\bar X_l^{(t)})\right).
\end{gathered}
\end{equation}
The final outputs $\{X_l^{(T)}\}_{l=1}^L$ across different levels are then flattened along the spatial dimensions and concatenated to derive the visual memory $\mathcal{M}\in\mathbb{R}^{L H_p W_p\times d}$.

\subsubsection{Box-guided Query Decoder} Our query decoder follows the design of Deformable DETR~\cite{zhu2021deformable}, but differs in the initialization of instance queries: instead of using static learned embeddings, our queries are instantiated directly from the VLM outputs. Consider a response containing $N$ predicted instances, each represented by a bounding box followed by an \texttt{<OBJ>} token. For the $i$-th instance, let $h_i$ be the hidden state of its \texttt{<OBJ>} token and $b_i=[x_{i1},y_{i1},x_{i2},y_{i2}]$ its normalized box. We initialize the content query as $q_i=W_h\,h_i$ and the positional query as $p_i=\operatorname{MLP}(\phi(b_i))$, where $\phi$ denotes Fourier encoding.

Our decoder consists of six deformable blocks, each containing a self-attention module, a multi-scale deformable cross-attention module, and a feed-forward network. In the self-attention module, the $N$ instance queries interact with one another, where $q_i+p_i$ serves as both the query and the key, and $q_i$ serves as the value. In cross-attention, each instance query aggregates visual features from the memory $\mathcal{M}$ through multi-scale deformable attention conditioned on its reference box $r_i=[\frac{x_{i1}+x_{i2}}{2},\frac{y_{i1}+y_{i2}}{2},x_{i2}-x_{i1},y_{i2}-y_{i1}]$.

\subsubsection{Instance-Conditioned Dense Prediction} Given the multi-level visual features from the visual feature adapter and the query embeddings from the box-guided query decoder, this stage fuses the visual features, predicts an initial mask for each instance, and refines it across multiple scales.

\textbf{Multi-level Pixel Fusion.} Deep features encode high-level semantic information, while shallow features preserve fine-grained spatial details. To combine visual information from different levels, we first fuse the adapted visual features $\{X_l^{(T)}\}_{l=1}^{L}$ from top to bottom:
\begin{equation}
G_l=
\begin{cases}
\operatorname{Conv}_{1\times1}\!\left(X_L^{(T)}\right),
& l=L,\\[4pt]
\operatorname{GELU}\!\left(
\operatorname{GroupNorm}\!\left(
\operatorname{Conv}_{3\times3}\!\left(
G_{l+1}+\operatorname{Conv}_{1\times1}\!\left(X_l^{(T)}\right)
\right)\right)\right),
& l=L-1,\ldots,1.
\end{cases}
\label{eq:multilevel_pixel_fusion}
\end{equation}
To recover color, texture, and edge details that may be lost in the fused visual feature, we additionally resize the input image $I$ to $1024\times1024$ and feed it into a two-layer CNN to extract a shallow feature map. Then, we upsample $G_1$ to the same resolution $512\times 512$ and combine the two features by element-wise addition:
\begin{equation}
\begin{aligned}
  P&=\operatorname{Up}(G_1)+\operatorname{CNN}(I).
\end{aligned}
\end{equation}

\textbf{Dynamic Mask Head.} Let $\widetilde q_i$ denote the query representation of the $i$-th instance derived from the box-guided query decoder. We first apply an MLP to $\widetilde q_i$ to generate an instance-aware dynamic kernel, which is then applied to $P$ to produce pixel-wise logits:
\begin{equation}
  [k_i,\beta_i]=\operatorname{MLP}(\widetilde q_i),
  \qquad
  M_i=k_i^{\top}P+\beta_i.
\end{equation}

\textbf{Coarse-to-fine Refinement.} A single mask prediction is often insufficient for thin structures and translucent boundaries. Starting from $M_i^{(0)}=M_i$, we progressively refine the mask through three upsampling stages ($j=1,2,3$). At stage $j$, we upsample the previous stage feature $F_i^{(j-1)}$ and concatenate it with the resized image $I$, pixel feature $P$, and previous mask predictions $C_i^{(j)}$:
\begin{equation}
  U_i^{(j)}=\operatorname{Conv}_{1\times1}\!\left(
    \operatorname{Up}(F_i^{(j-1)})
    \mathbin{\Vert}\operatorname{Interp}_j(I)
    \mathbin{\Vert}\operatorname{Interp}_j(P)
    \mathbin{\Vert}C_i^{(j)}
  \right),
\end{equation}
where $\operatorname{Up}$ denotes transposed convolution, $\operatorname{Interp}_j$ denotes bilinear interpolation to the resolution of stage $j$, and $\Vert$ denotes channel concatenation. We then modulate $U_i^{(j)}$ with the instance query $\widetilde q_i$ via channel-wise scale and shift, and pass it through residual convolutional blocks $f_{\mathrm{rc}}^{(j)}$ to predict a logit residual:
\begin{equation}
  [\gamma_i^{(j)},\delta_i^{(j)}]=\operatorname{MLP}_j(\widetilde q_i),
  \qquad
  F_i^{(j)}=f_{\mathrm{rc}}^{(j)}\!\left(\big(1+\operatorname{tanh}(\gamma_i^{(j)})\big)\odot U_i^{(j)}+\delta_i^{(j)}\right),
\end{equation}
\begin{equation}
  M_i^{(j)}=\operatorname{Interp}_j\big(M_i^{(j-1)}\big)+\operatorname{Conv}_{1\times1}\big(F_i^{(j)}\big).
\end{equation}
Finally, the output prediction is obtained by $A_i=\operatorname{Sigmoid}(M_i^{(3)})$.

%% file: sections/04_training.tex
\section{Training Pipeline}
\label{sec:training_pipeline}

We train \modelname{} with a three-stage Supervised Fine-Tuning (SFT)
curriculum followed by mixed-task Reinforcement Learning (RL). The curriculum
first establishes a shared output interface, then improves high-resolution and
dense perception, and finally connects autoregressive localization with
pixel-level prediction. After SFT, RL directly optimizes the quality of the
complete prediction rather than the likelihood of individual output tokens.
The data composition across the training pipeline is summarized in
Table~\ref{tab:training_data_composition}.

\begin{table*}[t]
  \centering
  \caption{Training data composition. The SFT and RL rings show task distributions. The segmentation warm-up and matting fine-tuning rings show data-source proportions: Public vs.\ WDE (\modelname Data Engine). Center numbers indicate training sample counts.}
  \label{tab:training_data_composition}
  \includegraphics{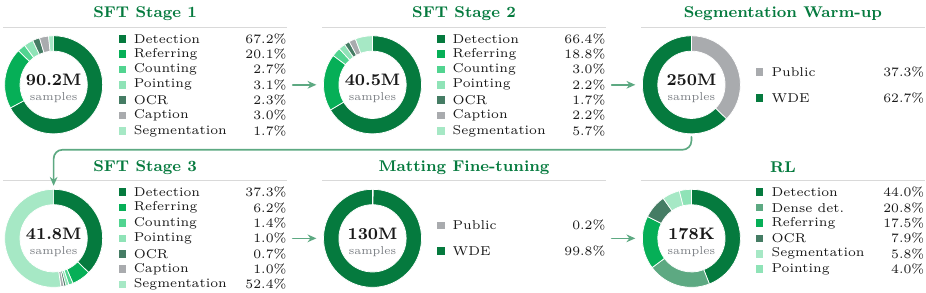}
\end{table*}

\subsection{Curriculum Supervised Fine-Tuning}

\textbf{Stage 1: Task Alignment.}
This stage adapts the model to the new tokens and task-specific structured outputs. Starting from Qwen3-VL with an extended vocabulary, we freeze the vision encoder and visual-language projector and train only the language model on 90M examples covering all target tasks for 13K optimization steps. The model learns to follow different task instructions, produce the structured outputs, and return \texttt{None} when the target is absent. For pixel-task examples, we impose only next-token prediction supervision (NTP) without activating the pixel decoder, thereby establishing an initial correspondence between each predicted instance and its \texttt{<OBJ>} token. In this stage, we use 16--512 visual tokens per image and limit the context length to 4,096 tokens. Notably, early training can be dominated by extremely dense samples (e.g., containing hundreds of boxes in a single response), making the model highly prone to repetitive output collapse. To mitigate this issue, we primarily use samples with at most 50 boxes or points per category during this stage.

\textbf{Stage 2: Resolution and Density Scaling.}
Building on Stage 1, we unfreeze the vision encoder, projector, and language model. This stage serves two purposes. First, we raise the maximum input resolution to 1,960 visual tokens, preserving the visual details for recognizing small objects. Second, we increase the maximum box/point number within one response to 500, exposing the model to crowded scenes for dense detection. The maximum context length is extended to 8,192 tokens. We train on approximately 40M examples for 9K optimization steps. During this stage, pixel-level examples remain supervised solely by NTP.

\textbf{Stage 3: Alternating Token--Pixel Optimization.}
This stage aims to equip the VLM with pixel-level perception by establishing the \texttt{<OBJ>} token as the bridge between the VLM and the lightweight pixel decoder. To initialize the pixel decoder before alternating token--pixel training, we first jointly train the SFT-S2 VLM and the randomly initialized decoder on 250M segmentation samples for 100K steps with the following segmentation loss function:
\begin{equation}
  \mathcal{L}_{\mathrm{seg}}
  = 20\mathcal{L}_{\mathrm{focal}}+\mathcal{L}_{\mathrm{dice}},
\end{equation}
where the focal term emphasizes hard pixels under foreground--background imbalance, while the Dice term directly optimizes mask overlap. After warm-up, we retain only the decoder weights and the learned \texttt{<OBJ>} token embedding. We attach them to the original Stage~2 checkpoint and jointly train the VLM and decoder for 9{,}000 steps. To preserve localization ability, we alternate NTP and pixel-level tasks at a 1:1 ratio, which can be expressed as the following loss function:
\begin{equation}
  \mathcal{L}_{\mathrm{joint}}
  =\mathcal{L}_{\mathrm{NTP}}
  +10 \mathcal{L}_{\mathrm{seg}}.
\end{equation}
After the alternating training, we specialize the pixel decoder for continuous alpha prediction while keeping the VLM frozen. We use 130M matting samples to further finetune the decoder for 100K steps. Following standard practice in image matting~\cite{guo2024incontextmatting}, we combine $L_1$, gradient, and Laplacian losses to supervise alpha estimation, and additionally employ $\mathcal{L}_{\mathrm{hard}}$ to emphasize difficult foreground and background pixels:
\begin{equation}
  \mathcal{L}_{\mathrm{mat}}
  =200\mathcal{L}_{l1}
  +5\mathcal{L}_{\mathrm{grad}}
  +50\mathcal{L}_{\mathrm{lap}}
  +\mathcal{L}_{\mathrm{hard}}.
\end{equation}

\subsection{Mixed-Task Reinforcement Learning}
Although SFT provides a strong initialization, its token-level likelihood
objective cannot directly optimize task rewards. We therefore perform
mixed-task reinforcement learning on an uncertainty-filtered corpus.

\textbf{Uncertainty-Aware Sample Selection.}
GRPO requires relative advantages to provide learning signals. Before RL training, we therefore use the SFT model for data sampling. For each candidate prompt $x$, we sample a group of responses and compute their corresponding task rewards. Let $\mu_x$ and $\sigma_x$ denote the mean and standard deviation of the group rewards, respectively. The uncertainty score for each sample is defined as follows:
\begin{equation}
  U(x)=4\mu_x(1-\mu_x)\sigma_x.
\end{equation}
This score favors moderately difficult samples with high reward variance, on which the model succeeds in some rollouts but fails in others. Such samples have high uncertainty and yield stronger relative-advantage signals, thereby improving the efficiency of RL training. Using this score, we select approximately 140K samples for RL training. We supplement this mixture with targeted box-supervised examples and pointing data, yielding 178K examples in total. Table~\ref{tab:training_data_composition} shows the complete task distribution. Data selection and reward details are provided in Appendices~\ref{sec:appendix_rl_datasets} and~\ref{sec:appendix_rl_rewards}, respectively.

\textbf{GRPO with Grounding Rewards.}
On the resulting training corpus, we perform GRPO using grounding rewards computed over complete structured responses. These rewards directly evaluate the grounding quality of the model predictions rather than individual tokens.
For each selected image--prompt pair $x$, we sample $G=8$ complete responses $\{o_i\}_{i=1}^{G}$ from the current policy and parse each response into a set of predicted objects. Let $\hat{\mathcal{B}}_i$ and $\mathcal{B}^{*}$ denote the predicted and ground-truth box sets, respectively. We perform one-to-one Hungarian matching between predicted and ground-truth boxes within each semantic class, and denote the set of matched pairs as $\mathcal{M}_i$. The grounding reward for response $o_i$ is defined as:
\begin{equation}
  r_i =
  \frac{2\sum_{(\hat b,b^{*})\in\mathcal{M}_i}
  \operatorname{IoU}(\hat b,b^{*})^{3}}
  {N_{\mathrm{gt}}+N_{\mathrm{pred}}}.
\end{equation}
This reward measures localization quality, while penalizing both missed and duplicate objects. Cubing the IoU makes the reward more sensitive to localization improvements at high IoU values, encouraging more precise box predictions. For segmentation and matting, RL uses only box-based rewards, without evaluating predicted masks or alpha mattes. For referring segmentation, a prediction is valid only when its box is followed by the required \texttt{<OBJ>} token. For negative samples, the reward is 1 only when the model predicts no object.

%% file: sections/05_experiments.tex
\section{Experiments}
\label{sec:experiments}

We evaluate \modelname{} across five object-level tasks (object detection, referring object detection, object counting, object pointing, and OCR) and two pixel-level tasks (referring segmentation and image matting). At inference, we use greedy decoding with temperature 0 and limit each image to 1,960 visual tokens.
We primarily compare with the Qwen series as general-purpose VLM baselines and with two representative VLM-based grounding models, Rex-Omni~\cite{jiang2026detect} and LocateAnything~\cite{wang2026locateanything}. For Qwen3-VL, we evaluate the Instruct variants; for Qwen3.5, we enable thinking mode because it performs better in our evaluation.

\subsection{Main Results}
\label{sec:main_results}

\subsubsection{Object Detection}

\textbf{Benchmarks.}
We evaluate common object detection on COCO~\cite{lin2014microsoft}, long-tailed object detection on LVIS~\cite{gupta2019lvis}, and dense and tiny-object detection on Dense200~\cite{jiang2026detect} and VisDrone~\cite{du2019visdrone}.

\textbf{Metrics.}
Average Precision (AP) relies on calibrated confidence scores and is therefore not directly applicable to generative VLMs. Following Rex-Omni~\cite{jiang2026detect} and LocateAnything~\cite{wang2026locateanything}, we report F1 scores. A predicted box is regarded as a true positive if its IoU with an unmatched ground-truth box of the same category exceeds a threshold. Precision and recall are computed per category and averaged, and F1 is their harmonic mean. We report F1 at IoU=0.5, at IoU=0.95, and averaged over IoU thresholds from 0.5 to 0.95 with a step of 0.05 (denoted as ``Mean'').

\begin{table*}[t]
  \centering
  \caption{Results on general object detection benchmarks COCO and LVIS, and dense object detection benchmarks Dense200 and VisDrone. COCO and LVIS are evaluated in a zero-shot setting. We report F1 at IoU=0.5, IoU=0.95, and averaged over IoU thresholds from 0.5 to 0.95 (Mean). \textbf{Bold} and \underline{underline} denote the best and second-best results among VLMs.}
  \label{tab:exp_detection}
  \ExpTableSetup
  \begin{tabularx}{\linewidth}{@{}l|*{3}{C}|*{3}{C}|*{3}{C}|*{3}{C}@{}}
    \toprule
    \multirow{2}{*}{Method} & \multicolumn{3}{c|}{COCO} & \multicolumn{3}{c|}{LVIS} & \multicolumn{3}{c|}{Dense200} & \multicolumn{3}{c}{VisDrone} \\
    \cmidrule{2-13}
     & 0.5 & 0.95 & Mean & 0.5 & 0.95 & Mean & 0.5 & 0.95 & Mean & 0.5 & 0.95 & Mean \\
    \midrule
    \rowcolor{gray!12}
    \multicolumn{13}{@{}l}{\textit{Specialized Detectors}} \\
    GroundingDINO-T~\cite{liu2024grounding} & 69.8 & 23.0 & 56.6 & 47.7 & 22.7 & 38.8 & 36.9 & 19.7 & 33.1 & 55.2 & 3.9 & 38.5 \\
    \midrule
    \rowcolor{gray!12}
    \multicolumn{13}{@{}l}{\textit{Vision-Language Models}} \\
    Qwen3-VL-2B~\cite{bai2025qwen3} & 62.9 & 13.1 & 45.2 & 58.5 & 17.8 & 40.8 & 14.7 & 1.8 & 10.0 & 37.0 & 1.5 & 23.1 \\
    Qwen3-VL-4B~\cite{bai2025qwen3} & 66.6 & 14.9 & 48.7 & 61.3 & 19.5 & 44.0 & 20.9 & 3.7 & 15.2 & 49.5 & 1.8 & 31.0 \\
    Qwen3-VL-8B~\cite{bai2025qwen3} & 65.9 & 14.8 & 47.8 & 59.7 & 19.6 & 42.9 & 19.5 & 2.9 & 13.8 & 47.7 & 2.0 & 29.5 \\
    Qwen3.5-4B~\cite{qwen3.5} & 67.6 & 12.9 & 48.2 & 61.1 & 17.8 & 42.2 & 28.4 & 2.3 & 18.9 & 50.3 & 1.3 & 29.9 \\
    Qwen3.5-9B~\cite{qwen3.5} & 68.9 & 13.8 & 49.5 & 62.2 & 19.4 & 43.7 & 26.2 & 2.5 & 18.0 & 51.0 & 1.4 & 31.3 \\
    \midrule
    Rex-Omni-3B~\cite{jiang2026detect} & 72.0 & 15.9 & 52.9 & 64.3 & 20.7 & 46.9 & 78.4 & 10.3 & 58.3 & 61.6 & 1.5 & 35.8 \\
    LocateAnything-3B~\cite{wang2026locateanything} & 70.1 & \underline{19.3} & 54.7 & 62.3 & \textbf{31.1} & 50.7 & 74.0 & \underline{18.5} & 58.7 & 63.0 & \underline{3.2} & 39.9 \\
    \midrule
    \oursrow \textbf{\modelname{}-2B}  & \underline{72.6} & 18.7 & \underline{55.2} & \underline{69.5} & 27.5 & \underline{53.3} & \underline{79.1} & 14.1 & \underline{61.4} & \underline{64.5} & 2.8 & \underline{40.4} \\
    \oursrow \textbf{\modelname{}-4B} & \textbf{74.0} & \textbf{19.5} & \textbf{57.1} & \textbf{71.1} & \underline{28.8} & \textbf{55.3} & \textbf{86.9} & \textbf{23.2} & \textbf{71.3} & \textbf{68.1} & \textbf{3.9} & \textbf{44.7} \\
    \bottomrule
  \end{tabularx}
\end{table*}

\textbf{Results.} As shown in Table~\ref{tab:exp_detection}, \modelname{} consistently outperforms VLMs at comparable model scales in mean F1 across general, long-tailed, and dense detection benchmarks. The gains are especially pronounced on Dense200 and VisDrone, where accurate recognition must be maintained over many small and crowded instances. Its strong performance under the strict IoU${}={}$0.95 criterion further indicates that the improvement reflects both broad object coverage and precise localization.

\subsubsection{Referring Object Detection}

\textbf{Benchmarks.}
We evaluate referring object detection on the public RefCOCOg~\cite{mao2016generation}, HumanRef~\cite{jiang2025referring}, Ref-Adv~\cite{dong2026refadv}, PR-Bench~\cite{gao2026motto}, DVGBench~\cite{zhou2026dvgbench}, and Ref-L4~\cite{chen2025refl4} benchmarks. We also include RefCOCOg-Rej~\cite{li2026groundingme}, in which each modified expression has no matching object in the image. Furthermore, we introduce WeRef-Bench, an in-house benchmark comprising 2,880 English and Chinese queries. It covers single-target referring, multi-target referring, and absent-target queries, providing more challenging and practical evaluations than the public positive, single-target benchmarks.

\textbf{Metrics.}
For most benchmarks, we report F1 Mean as the primary metric. For RefCOCOg-Rej, we report the Rejection Rate, where a prediction is correct only when no box is produced. For WeRef-Bench and PR-Bench, we report both F1 Mean and Rejection Rate to evaluate target-present and absent-target queries, respectively.

\begin{table*}[t]
  \centering
  \caption{Results on referring object detection. Entries are mean F1 over IoU thresholds from 0.5 to 0.95. Rej denotes the rejection rate (\%) on absent-target queries; the RefCOCOg Rej is evaluated on RefCOCOg-Rej. For Ref-Adv, we use the Ref-Adv-s split. \textbf{Bold} and \underline{underline} denote the best and second-best results among VLMs.}
  \label{tab:exp_rec}
  \ExpTableSetup
  \setlength{\tabcolsep}{1.8pt}
  \begin{tabularx}{\linewidth}{@{}l|*{3}{>{\centering\arraybackslash}p{.048\linewidth}}|*{2}{>{\centering\arraybackslash}p{.062\linewidth}}|*{2}{>{\centering\arraybackslash}p{.056\linewidth}}|>{\centering\arraybackslash}p{.100\linewidth}|>{\centering\arraybackslash}p{.080\linewidth}|C|C@{}}
    \toprule
    \multirow{2}{*}{Method} & \multicolumn{3}{c|}{RefCOCOg} & \multicolumn{2}{c|}{WeRef-Bench} & \multicolumn{2}{c|}{PR-Bench} & \multirow{2}{*}{HumanRef} & \multirow{2}{*}{Ref-Adv} & \multirow{2}{*}{DVG} & \multirow{2}{*}{Ref-L4} \\
    \cmidrule(lr){2-4} \cmidrule(lr){5-6} \cmidrule(lr){7-8}
     & \multicolumn{1}{c}{val}
     & \multicolumn{1}{c}{test}
     & \multicolumn{1}{c|}{Rej}
     & \multicolumn{1}{c}{Mean}
     & \multicolumn{1}{c|}{Rej}
     & \multicolumn{1}{c}{Mean}
     & \multicolumn{1}{c|}{Rej}
     & & & & \\
    \midrule
    \rowcolor{gray!12}
    \multicolumn{12}{@{}l}{\textit{Specialized Detectors}} \\
    GroundingDINO-T~\cite{liu2024grounding} & 45.9 & 46.8 & 0.0 & 21.1 & 14.7 & 25.2 & 0.0 & 25.2 & 19.7 & 24.1 & 23.1 \\
    \midrule
    \rowcolor{gray!12}
    \multicolumn{12}{@{}l}{\textit{Vision-Language Models}} \\
    Qwen3-VL-2B~\cite{bai2025qwen3} & 70.8 & 71.0 & 1.7 & 30.8 & 22.8 & 41.6 & 4.3 & 63.2 & 34.7 & 35.2 & 67.9 \\
    Qwen3-VL-4B~\cite{bai2025qwen3} & 74.1 & 74.3 & 18.8 & 46.3 & 61.9 & 49.9 & 10.8 & 69.4 & 44.4 & 42.4 & 72.5 \\
    Qwen3-VL-8B~\cite{bai2025qwen3} & 74.6 & 74.9 & 22.5 & 36.7 & 18.3 & 50.3 & 7.1 & 71.6 & 45.0 & \underline{43.4} & 72.4 \\
    Qwen3.5-4B~\cite{qwen3.5} & 73.3 & 73.5 & 18.9 & 48.1 & 71.6 & 49.6 & 8.7 & 75.8 & 44.2 & 40.2 & 70.4 \\
    Qwen3.5-9B~\cite{qwen3.5} & 74.3 & 74.6 & 15.2 & 50.7 & 70.7 & \underline{50.8} & 8.3 & 77.8 & \underline{47.3} & 42.5 & 72.0 \\
    \midrule
    Rex-Omni-3B~\cite{jiang2026detect} & 73.6 & 74.3 & 0.5 & 50.3 & 43.3 & 42.7 & 0.2 & \underline{79.9} & 40.9 & 33.2 & 65.2 \\
    LocateAnything-3B~\cite{wang2026locateanything} & 76.7 & 77.6 & 0.6 & 51.7 & 73.0 & 44.5 & 0.2 & 78.7 & 43.5 & 34.7 & 74.3 \\
    \midrule
    \oursrow \textbf{\modelname{}-2B}  & \underline{77.1} & \underline{78.2} & \underline{50.2} & \underline{73.2} & \underline{92.9} & 49.8 & \underline{37.0} & 77.4 & 43.6 & 40.8 & \underline{75.1} \\
    \oursrow \textbf{\modelname{}-4B} & \textbf{80.8} & \textbf{81.8} & \textbf{67.4} & \textbf{78.7} & \textbf{96.4} & \textbf{54.1} & \textbf{52.6} & \textbf{80.7} & \textbf{50.5} & \textbf{45.2} & \textbf{78.1} \\
    \bottomrule
  \end{tabularx}
\end{table*}

\textbf{Results.} Table~\ref{tab:exp_rec} shows that \modelname{} consistently outperforms existing VLMs across single-target, multi-target, and absent-target settings. These results demonstrate that \modelname{} localizes referred objects more accurately and rejects absent-target queries more reliably.

\subsubsection{Object Counting}

\textbf{Benchmarks.}
We evaluate object counting on CountQA~\cite{tamarapalli2025countqa}, PixMo-Count~\cite{deitke2024molmo}, CLOC~\cite{lei2026countanything}, KubriCount~\cite{liu2026countanythinggranularity}, and MixCount~\cite{dumery2026mixcount}. CountQA and PixMo-Count focus on counting in natural images, while CLOC evaluates cross-domain generalization. KubriCount stresses multi-grained target specification with controlled distractors, and MixCount targets fine-grained open-vocabulary counting in mixed-object scenes.

\textbf{Metrics.}
For models evaluated through detection output, we use the number of predicted target boxes as the count; for models that return a count directly, we extract the numerical answer from the response. We report exact-match accuracy: the percentage of queries for which the predicted count equals the ground-truth count. Thus, an error of even one object is counted as incorrect, and higher values are better.

\begin{table*}[t]
  \centering
  \caption{Object-counting exact-match accuracy (\%, $\uparrow$). \textbf{Bold} and \underline{underline} denote the best and second-best results among VLM-based models, respectively; the specialized detector is excluded from this ranking.}
  \label{tab:exp_counting}
  \ExpTableSetup
  \setlength{\tabcolsep}{2.0pt}
  \begin{tabularx}{\linewidth}{@{}l|>{\centering\arraybackslash}p{0.09\linewidth}|*{2}{C}|*{2}{C}|*{2}{C}|*{2}{C}@{}}
    \toprule
    \multirow{2}{*}{Method} & \multirow{2}{*}{CountQA} &
    \multicolumn{2}{c|}{PixMo-Count} & \multicolumn{2}{c|}{CLOC} &
    \multicolumn{2}{c|}{KubriCount} & \multicolumn{2}{c}{MixCount} \\
    \cmidrule{3-10}
    & & val & test & val & test & testA & testB & val & test \\
    \midrule
    \rowcolor{gray!12}
    \multicolumn{10}{@{}l}{\textit{Specialized Detectors}} \\
    Grounding DINO-T~\cite{liu2024grounding} & 9.6 & 44.8 & 48.3 & 12.4 & 9.5 & 5.2 & 6.5 & 1.7 & 2.8 \\
    \midrule
    \rowcolor{gray!12}
    \multicolumn{10}{@{}l}{\textit{Vision-Language Models}} \\
    Qwen3-VL-2B~\cite{bai2025qwen3} & 20.4 & 61.3 & 60.4 & 16.3 & 15.3 & 4.3 & 7.8 & 1.8 & 2.7 \\
    Qwen3-VL-4B~\cite{bai2025qwen3} & 29.2 & 69.6 & 65.6 & 19.7 & 18.6 & 7.1 & 12.1 & 3.5 & 4.3 \\
    Qwen3-VL-8B~\cite{bai2025qwen3} & 22.4 & 68.2 & 59.4 & 18.5 & 17.2 & 6.6 & 10.8 & 3.2 & 3.5 \\
    Qwen3.5-4B~\cite{qwen3.5} & 40.0 & 80.6 & 76.9 & 21.1 & 21.7 & 10.0 & 16.2 & 3.9 & 5.2 \\
    Qwen3.5-9B~\cite{qwen3.5} & 41.6 & 82.6 & 80.4 & 23.1 & 22.4 & 11.9 & 19.1 & 5.8 & 7.0 \\
    \midrule
    Rex-Omni-3B~\cite{jiang2026detect} & 37.6 & 82.2 & 71.9 & 19.0 & 22.3 & 13.5 & 18.8 & 9.9 & 11.4 \\
    LocateAnything-3B~\cite{wang2026locateanything} & \underline{52.4} & 88.7 & 82.8 & 22.8 & 23.9 & 18.5 & 22.4 & 12.5 & 15.2 \\
    \midrule
    \oursrow \textbf{\modelname{}-2B}  & 47.9 & \underline{90.9} & \underline{88.1} & \underline{27.5} & \underline{26.9} & \underline{29.5} & \underline{32.4} & \underline{30.1} & \underline{32.2} \\
    \oursrow \textbf{\modelname{}-4B} & \textbf{57.1} & \textbf{93.7} & \textbf{89.6} & \textbf{28.2} & \textbf{28.3} & \textbf{37.7} & \textbf{40.1} & \textbf{42.0} & \textbf{40.4} \\
    \bottomrule
  \end{tabularx}
\end{table*}

\textbf{Results.} Table~\ref{tab:exp_counting} shows that \modelname{} achieves the highest exact-match accuracy on every reported benchmark and split. It reaches 93.7\%/89.6\% on PixMo-Count validation/test and 42.0\%/40.4\% on MixCount validation/test, indicating accurate counts in both natural images and fine-grained mixed-object scenes.

\subsubsection{Object Pointing}

\textbf{Benchmarks.}
We evaluate object pointing on two dedicated pointing benchmarks, PointBench~\cite{cheng2025pointarena} and PixMo-Points~\cite{deitke2024molmo}. We also evaluate pointing on the detection benchmarks COCO, LVIS, Dense200, and VisDrone, following the protocol of Rex-Omni~\cite{jiang2026detect}.

\textbf{Metrics.}
A predicted point is correct when it falls inside the corresponding ground-truth mask (or the annotated target box where masks are unavailable). Each target contributes at most one true positive, so duplicate points reduce precision and missed targets reduce recall. We report F1@Point as the harmonic mean of precision and recall. For detection-derived benchmarks, precision and recall are first averaged across categories. 

\begin{table*}[t]
  \centering
  \caption{Results on object pointing. \textbf{Bold} and \underline{underline} denote the best and second-best reported results, respectively.}
  \label{tab:exp_pointing}
  \ExpTableSetup
  \setlength{\tabcolsep}{2.5pt}
  \begin{tabularx}{\linewidth}{@{}l|>{\centering\arraybackslash}p{.11\linewidth}>{\centering\arraybackslash}p{.14\linewidth}|*{4}{C}@{}}
    \toprule
    Method & PointBench & PixMo-Points & COCO & LVIS & Dense200 & VisDrone\\
    \midrule
    Qwen3-VL-2B~\cite{bai2025qwen3} & 61.2 & 62.1 & 72.7 & 61.9 & 29.5 & 35.7\\
    Qwen3-VL-4B~\cite{bai2025qwen3} & 57.1 & 71.3  & 74.5 & 65.8 & 37.7 & 45.6\\
    Qwen3-VL-8B~\cite{bai2025qwen3} & 62.5 & 72.2  & 77.5 & 67.7 & 42.2 & 46.4\\
    Qwen3.5-4B~\cite{qwen3.5} & 61.5 & 59.2  & 72.7 & 62.4 & 7.4 & 38.3\\
    Qwen3.5-9B~\cite{qwen3.5} & 66.4 & 62.6  & 74.4 & 64.8 & 10.5 & 40.6\\
    \midrule
    Rex-Omni-3B~\cite{jiang2026detect} & \underline{69.9} & 72.9  & 80.5 & 70.8 & 82.5 & 58.9\\
    LocateAnything-3B~\cite{wang2026locateanything} & 56.9 & 68.6  & 83.9 & 76.6 & \textbf{87.6} & 60.4\\
    \midrule
    \oursrow \textbf{\modelname{}-2B}  & 62.9 & \underline{73.2} & \textbf{85.8} & \underline{78.9} & 83.2 & \underline{62.4} \\
    \oursrow \textbf{\modelname{}-4B} & \textbf{71.9} & \textbf{74.7}  & \underline{85.4} & \textbf{79.5} & \underline{87.0} & \textbf{64.4}\\
    \bottomrule
  \end{tabularx}
\end{table*}

\textbf{Results.} As shown in Table~\ref{tab:exp_pointing}, \modelname{} achieves the best results on both PointBench and PixMo-Points, demonstrating its ability to point directly to target objects. On the detection-derived benchmarks, it also performs strongly across general detection benchmarks, showing that this ability extends to diverse scenes.

\subsubsection{OCR}

\textbf{Benchmarks.}
We evaluate text detection and recognition on HierText~\cite{long2022towards}, ICDAR2015~\cite{karatzas2015icdar}, TotalText~\cite{ch2017total}, and SROIE~\cite{huang2019icdar2019}. These benchmarks cover dense document text, incidental scene text, curved text, and structured receipts.

\textbf{Metrics.}
We formulate OCR as a grounding task in which the recognized transcription serves as the category label. A prediction is considered correct only when its text exactly matches the ground-truth transcription and its box satisfies the IoU threshold. We then compute precision, recall, and F1 using the same matching procedure as object detection, and report F1 at IoU=0.5, at IoU=0.95, and averaged over thresholds from 0.5 to 0.95.

\begin{table*}[t]
  \centering
  \caption{Results on OCR (text grounding and recognition). We report F1 at IoU=0.5, IoU=0.95, and averaged over IoU thresholds from 0.5 to 0.95 (Mean). \textbf{Bold} and \underline{underline} denote the best and second-best results among VLM-based models, respectively; the specialist OCR system is excluded from this ranking.}
  \label{tab:exp_ocr}
  \ExpTableSetup
  \begin{tabularx}{\linewidth}{@{}l|*{3}{C}|*{3}{C}|*{3}{C}|*{3}{C}@{}}
    \toprule
    \multirow{2}{*}{Method} & \multicolumn{3}{c|}{HierText} & \multicolumn{3}{c|}{ICDAR2015} & \multicolumn{3}{c|}{TotalText} & \multicolumn{3}{c}{SROIE} \\
    \cmidrule{2-13}
     & 0.5 & 0.95 & Mean & 0.5 & 0.95 & Mean & 0.5 & 0.95 & Mean & 0.5 & 0.95 & Mean \\
    \midrule
    \rowcolor{gray!12}
    \multicolumn{13}{@{}l}{\textit{Specialist OCR Systems}} \\
    PP-OCRv5~\cite{cui2025paddleocr} & 45.2 & 3.4 & 30.5 & 38.2 & 1.2 & 25.6 & 40.2 & 0.7 & 25.7 & 77.7 & 5.6 & 58.6 \\
    \midrule
    \rowcolor{gray!12}
    \multicolumn{13}{@{}l}{\textit{Vision-Language Models}} \\
        Qwen3-VL-2B~\cite{bai2025qwen3} & 21.4 & 0.3 & 11.0 & 15.7 & 0.0 & 8.5 & 32.6 & 1.1 & 20.5 & 40.2 & 0.2 & 18.6 \\
    Qwen3-VL-4B~\cite{bai2025qwen3} & 39.9 & 0.9 & 22.7 & 52.0 & 0.8 & 30.0 & 48.3 & 3.3 & 33.0 & 69.0 & 1.0 & 40.4 \\
    Qwen3-VL-8B~\cite{bai2025qwen3} & 45.3 & 0.9 & 25.7 & 45.8 & 0.3 & 27.1 & 53.2 & 3.2 & 35.4 & 75.8 & 1.1 & 43.3 \\
    Qwen3.5-4B~\cite{qwen3.5} & 14.4 & 0.4 & 8.0 & 17.4 & 0.1 & 7.8 & 28.8 & 0.6 & 16.6 & 14.7 & 0.0 & 7.1 \\
    Qwen3.5-9B~\cite{qwen3.5} & 17.5 & 0.5 & 10.0 & 28.1 & 0.0 & 13.4 & 49.9 & 1.6 & 30.8 & 31.8 & 0.4 & 16.7 \\
    \midrule
    Rex-Omni-3B~\cite{jiang2026detect} & \underline{45.9} & 1.4 & 28.0 & 45.2 & 0.3 & 28.1 & 56.6 & 3.9 & 40.6 & 72.0 & 1.5 & 44.8 \\
    LocateAnything-3B~\cite{wang2026locateanything} & 39.2 & \underline{4.1} & 29.4 & 37.8 & 0.3 & 24.1 & 58.9 & \underline{5.1} & \underline{43.3} & 52.8 & 3.4 & 40.3 \\
    \midrule
    \oursrow \textbf{\modelname{}-2B}  & 44.3 & 2.5 & \underline{30.5} & \underline{65.3} & \underline{2.0} & \underline{42.6} & \underline{61.5} & 2.7 & 42.5 & \underline{85.3} & \underline{4.4} & \underline{62.1} \\
    \oursrow \textbf{\modelname{}-4B} & \textbf{51.6} & \textbf{5.3} & \textbf{39.0} & \textbf{69.1} & \textbf{2.3} & \textbf{47.8} & \textbf{70.1} & \textbf{6.9} & \textbf{52.0} & \textbf{87.4} & \textbf{7.3} & \textbf{68.1} \\
    \bottomrule
  \end{tabularx}
\end{table*}

\textbf{Results.} As shown in Table~\ref{tab:exp_ocr}, \modelname{} consistently outperforms both specialist OCR systems and general-purpose VLMs across all evaluated benchmarks. Its strong performance across dense documents, incidental and curved scene text, and structured receipts demonstrates robust end-to-end text perception under diverse layouts and visual conditions.

\subsubsection{Referring Segmentation}

Given an image and a natural-language expression, referring segmentation predicts a binary mask of the described objects, extending localization from box-level predictions to pixel-level dense outputs.

\textbf{Benchmarks.}
We evaluate referring segmentation on RefCOCO and RefCOCO+~\cite{yu2016modeling}, RefCOCOg~\cite{mao2016generation}, ReasonSeg~\cite{lai2024lisa} and Ref-ZOM~\cite{hu2023beyond}. ReasonSeg uses implicit queries that require visual reasoning and world knowledge. Ref-ZOM further includes expressions that refer to zero, one, or multiple targets.

\textbf{Metrics.}
For the RefCOCO family and Ref-ZOM, we report cumulative intersection over union (cIoU) over the entire evaluation set, computed as the total intersection between predicted and ground-truth masks divided by their total union across all samples. For ReasonSeg, we report generalized IoU (gIoU) on the validation and test splits. For Ref-ZOM, which contains negative expressions that refer to no target, we additionally report the rejection rate (Rej).

\begin{table*}[t]
  \centering
  \caption{Referring segmentation results. RefCOCO-family scores are cIoU, while ReasonSeg scores are gIoU. Ref-ZOM baseline scores are obtained using official checkpoints. A hyphen (-) denotes an unreported result. \textbf{Bold} and \underline{underline} denote the best and second-best reported results in each column, respectively.}
  \label{tab:exp_refseg}
  \ExpTableSetup
  \setlength{\tabcolsep}{1.2pt}
  \renewcommand{\arraystretch}{1.05}
  \begin{tabularx}{\linewidth}{@{}l|*{3}{C}|*{3}{C}|*{2}{C}|*{2}{C}|*{2}{C}@{}}
    \toprule
    \multirow{2}{*}{Method} & \multicolumn{3}{c|}{RefCOCO} & \multicolumn{3}{c|}{RefCOCO+} & \multicolumn{2}{c|}{RefCOCOg} & \multicolumn{2}{c|}{ReasonSeg} & \multicolumn{2}{c}{Ref-ZOM} \\
    \cmidrule(lr){2-4} \cmidrule(lr){5-7} \cmidrule(lr){8-9} \cmidrule(lr){10-11} \cmidrule(lr){12-13}
    & val & testA & testB & val & testA & testB & val & test & val & test & cIoU & Rej \\
    \midrule
    LISA-13B (ft)~\cite{lai2024lisa} & 76.0 & 78.8 & 72.9 & 65.0 & 70.2 & 58.1 & 69.5 & 70.5 & \underline{65.0} & \underline{61.3} & - & - \\
    GSVA-7B~\cite{xia2024gsva} & 77.2 & 78.9 & 73.5 & 65.9 & 69.6 & 59.8 & 72.7 & 73.3 & 50.5 & - & 66.6 & 41.9 \\
    LaSagnA-7B~\cite{wei2024lasagna} & 76.8 & 78.7 & 73.8 & 66.4 & 70.6 & 60.1 & 70.6 & 71.9 & - & - & 57.4 & 0.7 \\
    Text4Seg-7B~\cite{lan2024text4seg} & 79.3 & 81.9 & 76.2 & 72.1 & 77.6 & 66.1 & 72.1 & 73.9 & - & - & 54.5 & 25.9 \\
    VRS-HQ-7B~\cite{gong2025vrshq} & 73.5 & 77.5 & 69.5 & 61.7 & 67.6 & 54.3 & 66.7 & 67.5 & 55.2 & 51.7 & 63.0 & 13.9 \\
    UniPixel-7B~\cite{liu2026unipixel} & \underline{80.8} & \underline{83.0} & 77.4 & 75.3 & \underline{80.1} & 70.0 & 76.4 & \underline{77.1} & 60.5 & - & 62.9 & 0.3 \\
    Youtu-VL-4B~\cite{wei2026youtu} & 80.7 & 82.0 & \underline{78.4} & \underline{76.2} & 79.6 & \underline{71.4} & \underline{76.5} & 76.6 & - & - & 61.0 & 1.3 \\
    \midrule
    \oursrow \textbf{\modelname{}-2B}  & 79.7 & 82.9 & 78.1 & 73.1 & 79.0 & 66.3 & 74.4 & 74.9 & 60.9 & 56.9 & \textbf{73.4} & \underline{62.4} \\
    \oursrow \textbf{\modelname{}-4B} & \textbf{81.4} & \textbf{84.5} & \textbf{80.1} & \textbf{76.5} & \textbf{81.1} & \textbf{72.8} & \textbf{77.7} & \textbf{77.8} & \textbf{71.2} & \textbf{66.8} & \underline{72.2} & \textbf{86.0} \\
    \bottomrule
  \end{tabularx}
\end{table*}

\textbf{Results.} As shown in Table~\ref{tab:exp_refseg}, \modelname{} improves over all baselines across every RefCOCO-family split. It also demonstrates strong potential for reasoning segmentation even without explicit chain-of-thought (CoT) reasoning. Furthermore, the results on Ref-ZOM reveal that existing segmentation models tend to output a mask even when the target is absent, lacking the ability to reject negative expressions. In contrast, \modelname{} can not only output accurate masks when the target is present, but also correctly reject the expression when it is absent.

\subsubsection{Promptable Matting}
Matting predicts a continuous alpha matte that preserves fine structures and partial transparency, such as hair and fur. Existing matting evaluations typically assume a salient foreground~\cite{ke2022modnet,li2021privacy} or provide a ground-truth mask to specify the target~\cite{yang2025matanyone,shen2026sam2matting}. To evaluate promptable matting in a more practical setting, we use boxes, category names, and referring expressions to specify the target object.

\textbf{Benchmarks and Baselines.}
For box prompts, we use P3M-500-NP~\cite{li2021privacy}, AM-2K~\cite{li2022bridging}, and PPM-100~\cite{ke2022modnet}. For text prompts, we use our in-house benchmark WeRefMat. For the three established benchmarks, all methods use the same ground-truth box as prompt, and all baselines are run from their official checkpoints. We evaluate three SAM2-Matting~\cite{shen2026sam2matting} variants based on SAM2.1~\cite{ravi2025sam2} and SAM3~\cite{carion2026sam3}. For box prompts, SAM2-Matting first uses SAM2.1 or SAM3 to predict a mask from the ground-truth box and then performs matting, while \modelname{} constructs a sequence prefix from the ground-truth box, generates the subsequent \texttt{<OBJ>} token, and passes its hidden state to the pixel decoder to predict the alpha matte. On WeRefMat, we compare with SAM3's native text grounding and two MLLM-based localization settings, in which Qwen3-VL-8B-Instruct or Qwen3.5-9B first localizes the target and passes the resulting box to SAM2-Matting.

\textbf{Metrics.}
We report mean absolute difference (MAD), mean squared error (MSE), gradient error (Grad), connectivity error (Conn), and sum of absolute differences (SAD) between the predicted and ground-truth alpha mattes. Following common practice, MAD and MSE are multiplied by $10^3$. Lower values indicate better performance for all five metrics.

\begin{table*}[t]
  \centering
  \caption{Image matting results under box and language prompts. MAD and MSE are multiplied by $10^3$. On WeRefMat, all SAM2-Matting results use the SAM3 variant. \textbf{Bold} and \underline{underline} denote the best and second-best results, respectively.}
  \label{tab:exp_matting}
  \ExpTableSetup
  \setlength{\tabcolsep}{2.3pt}
  \begin{tabularx}{\linewidth}{@{}c|c|l|*{5}{C}@{}}
    \toprule
    Dataset & Prompt & Method & MAD$\downarrow$ & MSE$\downarrow$ & Grad$\downarrow$ & Conn$\downarrow$ & SAD$\downarrow$ \\
    \midrule
    \multirow{6}{*}{P3M-500-NP} & \multirow{6}{*}{Box} & Matte Anything~\cite{yao2024matte} & 24.4 & 20.4 & 20.9 & 40.4 & 40.9 \\
    & & SAM2-Matting (SAM2.1-T)~\cite{shen2026sam2matting} & 22.7 & 18.2 & 20.0 & 37.8 & 38.1 \\
    & & SAM2-Matting (SAM2.1-B+)~\cite{shen2026sam2matting} & 19.0 & 15.0 & 18.3 & 31.0 & 31.4 \\
    & & SAM2-Matting (SAM3)~\cite{shen2026sam2matting} & 15.4 & 11.4 & 17.6 & 25.1 & 25.8 \\
    & & \cellcolor{OursRowColor}\textbf{\modelname{}-2B} & \cellcolor{OursRowColor}\textbf{6.6} & \cellcolor{OursRowColor}\textbf{3.1} & \cellcolor{OursRowColor}\underline{10.7} & \cellcolor{OursRowColor}\textbf{10.6} & \cellcolor{OursRowColor}\textbf{11.2} \\
    & & \cellcolor{OursRowColor}\textbf{\modelname{}-4B} & \cellcolor{OursRowColor}\underline{7.3} & \cellcolor{OursRowColor}\underline{3.8} & \cellcolor{OursRowColor}\textbf{10.5} & \cellcolor{OursRowColor}\underline{11.7} & \cellcolor{OursRowColor}\underline{12.6} \\
    \midrule
    \multirow{6}{*}{AM-2K} & \multirow{6}{*}{Box} & Matte Anything~\cite{yao2024matte} & 13.2 & 8.7 & 13.9 & 22.0 & 23.1 \\
    & & SAM2-Matting (SAM2.1-T)~\cite{shen2026sam2matting} & 11.0 & 6.2 & 21.7 & 18.4 & 18.8 \\
    & & SAM2-Matting (SAM2.1-B+)~\cite{shen2026sam2matting} & 11.1 & 6.3 & 20.0 & 18.8 & 19.0 \\
    & & SAM2-Matting (SAM3)~\cite{shen2026sam2matting} & 12.1 & 7.5 & 19.1 & 19.5 & 20.1 \\
    & & \cellcolor{OursRowColor}\textbf{\modelname{}-2B} & \cellcolor{OursRowColor}\underline{5.7} & \cellcolor{OursRowColor}\underline{2.2} & \cellcolor{OursRowColor}\underline{8.1} & \cellcolor{OursRowColor}\underline{8.9} & \cellcolor{OursRowColor}\underline{9.7} \\
    & & \cellcolor{OursRowColor}\textbf{\modelname{}-4B} & \cellcolor{OursRowColor}\textbf{5.0} & \cellcolor{OursRowColor}\textbf{1.6} & \cellcolor{OursRowColor}\textbf{7.2} & \cellcolor{OursRowColor}\textbf{7.5} & \cellcolor{OursRowColor}\textbf{8.5} \\
    \midrule
    \multirow{6}{*}{PPM-100} & \multirow{6}{*}{Box} & Matte Anything~\cite{yao2024matte} & 10.5 & 6.7 & 66.3 & 100.1 & 102.4 \\
    & & SAM2-Matting (SAM2.1-T)~\cite{shen2026sam2matting} & 9.7 & 5.9 & 68.6 & 95.3 & 97.1 \\
    & & SAM2-Matting (SAM2.1-B+)~\cite{shen2026sam2matting} & 9.3 & 5.7 & 64.0 & 90.3 & 92.9 \\
    & & SAM2-Matting (SAM3)~\cite{shen2026sam2matting} & 10.8 & 6.9 & 65.9 & 100.6 & 103.2 \\
    & & \cellcolor{OursRowColor}\textbf{\modelname{}-2B} & \cellcolor{OursRowColor}\textbf{5.3} & \cellcolor{OursRowColor}\textbf{1.8} & \cellcolor{OursRowColor}\textbf{48.4} & \cellcolor{OursRowColor}\textbf{45.5} & \cellcolor{OursRowColor}\textbf{49.3} \\
    & & \cellcolor{OursRowColor}\textbf{\modelname{}-4B} & \cellcolor{OursRowColor}\underline{6.1} & \cellcolor{OursRowColor}\underline{2.4} & \cellcolor{OursRowColor}\underline{48.5} & \cellcolor{OursRowColor}\underline{53.2} & \cellcolor{OursRowColor}\underline{57.2} \\
    \midrule
    \multirow{10}{*}{WeRefMat} & \multirow{5}{*}{Category} & SAM2-Matting (SAM3)~\cite{shen2026sam2matting} & 70.8 & 64.9 & 45.0 & 245.6 & 241.6 \\
    & & Qwen3-VL-8B + SAM2-Matting & 80.7 & 74.4 & 53.6 & 270.5 & 267.1 \\
    & & Qwen3.5-9B + SAM2-Matting & 77.8 & 71.4 & 51.9 & 265.4 & 261.8 \\
    \cmidrule(lr){3-8}
    & & \cellcolor{OursRowColor}\textbf{\modelname{}-2B} & \cellcolor{OursRowColor}\underline{17.7} & \cellcolor{OursRowColor}\underline{13.4} & \cellcolor{OursRowColor}\underline{35.6} & \cellcolor{OursRowColor}\underline{63.6} & \cellcolor{OursRowColor}\underline{65.7} \\
    & & \cellcolor{OursRowColor}\textbf{\modelname{}-4B} & \cellcolor{OursRowColor}\textbf{16.4} & \cellcolor{OursRowColor}\textbf{12.5} & \cellcolor{OursRowColor}\textbf{35.0} & \cellcolor{OursRowColor}\textbf{60.3} & \cellcolor{OursRowColor}\textbf{62.0} \\
    \cmidrule(lr){2-8}
    & \multirow{5}{*}{Referring} & SAM2-Matting (SAM3)~\cite{shen2026sam2matting} & 51.9 & 50.3 & 34.3 & 162.1 & 162.2 \\
    & & Qwen3-VL-8B + SAM2-Matting & 23.0 & 21.1 & 22.6 & 78.3 & 78.1 \\
    & & Qwen3.5-9B + SAM2-Matting & 22.4 & 20.6 & 22.1 & 82.5 & 82.3 \\
    \cmidrule(lr){3-8}
    & & \cellcolor{OursRowColor}\textbf{\modelname{}-2B} & \cellcolor{OursRowColor}\underline{15.8} & \cellcolor{OursRowColor}\underline{14.2} & \cellcolor{OursRowColor}\underline{19.4} & \cellcolor{OursRowColor}\underline{65.3} & \cellcolor{OursRowColor}\underline{66.2} \\
    & & \cellcolor{OursRowColor}\textbf{\modelname{}-4B} & \cellcolor{OursRowColor}\textbf{9.8} & \cellcolor{OursRowColor}\textbf{8.4} & \cellcolor{OursRowColor}\textbf{15.9} & \cellcolor{OursRowColor}\textbf{37.8} & \cellcolor{OursRowColor}\textbf{39.0} \\
    \bottomrule
  \end{tabularx}
\end{table*}

\textbf{Results.} As shown in Table~\ref{tab:exp_matting}, \modelname{} is competitive with task-specific matting models under box prompts and consistently improves over the SAM2-Matting baseline. On WeRefMat, it performs strongly with both category and referring prompts, showing that language-conditioned target selection and boundary-sensitive transparency estimation can be handled effectively within a unified model.

%% file: sections/06_visualization.tex
\section{Qualitative Analysis}
\label{sec:qualitative}

We present qualitative comparisons between \modelname{} and representative baselines in Figures~\ref{fig:qual_detection}--\ref{fig:qual_matting_text}. Each figure is organized by task, and the selected examples below are discussed from top to bottom.

\begingroup
\captionsetup{hypcap=false,skip=0pt}
\newsavebox{\qualanalysis}
\newlength{\qualimageheight}
\newcommand{\qualocrstrip}[1]{%
  \includegraphics[width=\linewidth,bb=0 0 2085 2662,trim=#1,clip]{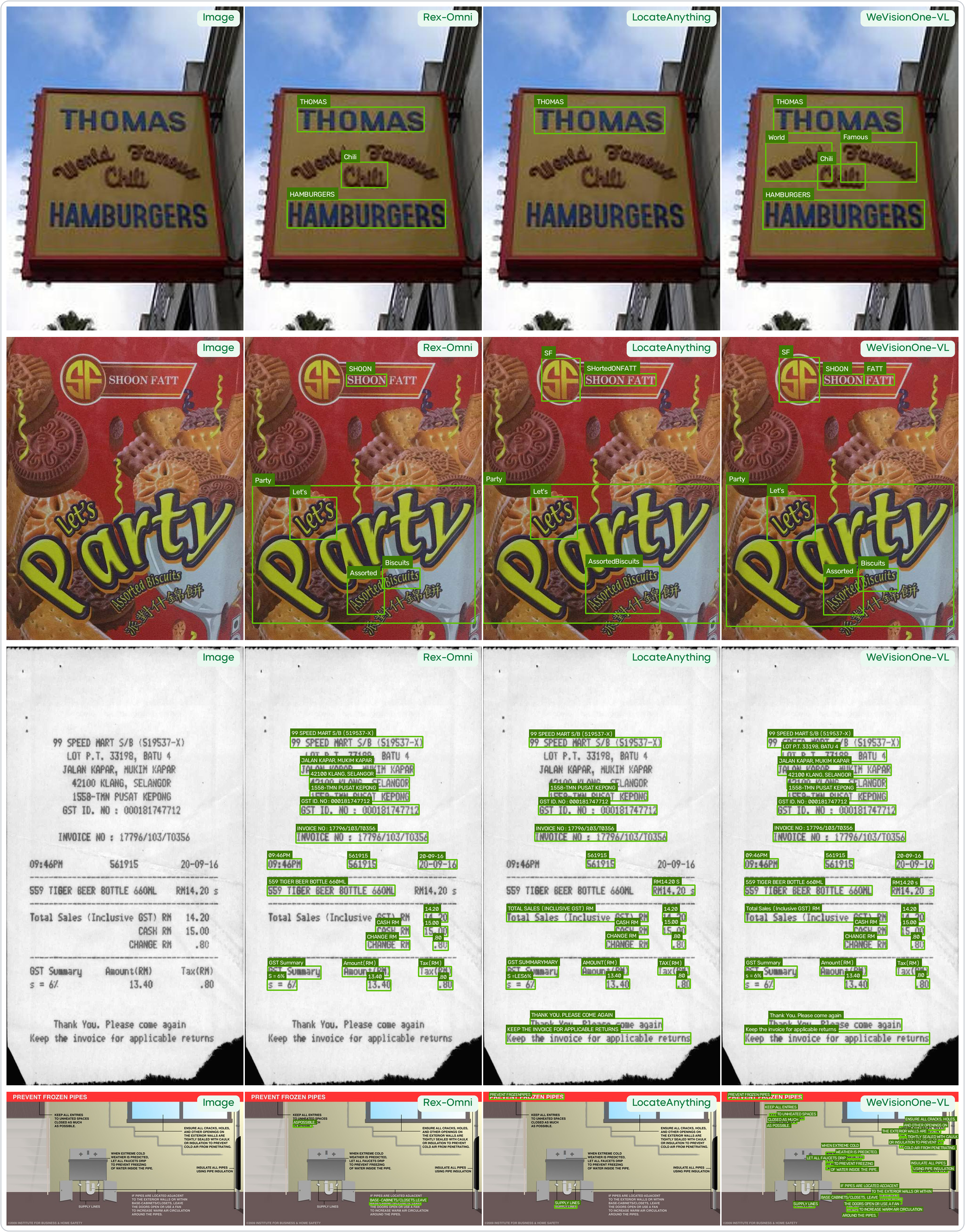}\par}

\par
\Needspace*{0.7\textheight}
\noindent\begin{minipage}{\linewidth}  
  \begin{lrbox}{\qualanalysis}
    \begin{minipage}{\linewidth}
  \captionof{figure}{Qualitative comparison on general and dense object detection.}
  \label{fig:qual_detection}
  \vspace{9pt}
  
\textbf{Object detection.}
Figure~\ref{fig:qual_detection} covers degraded visibility, adjacent instances, and dense scenes. In the first example, both baselines detect only the foreground elephant, while \modelname{} also recovers the distant elephants partly obscured by dust. Across the carrot and dense fruit examples, the baselines tend to merge adjacent instances into coarse-grained boxes, whereas \modelname{} more often preserves instance-level localization.
    \end{minipage}
  \end{lrbox}
  \setlength{\qualimageheight}{\dimexpr\textheight-\pagetotal-\ht\qualanalysis-\dp\qualanalysis-36pt\relax}
  \vspace*{12pt}
  \centering
  \begin{adjustbox}{max totalheight=\qualimageheight,max width=\linewidth}
    \begin{minipage}{\linewidth}
      \setlength{\parskip}{0pt}
      \includegraphics[width=\linewidth]{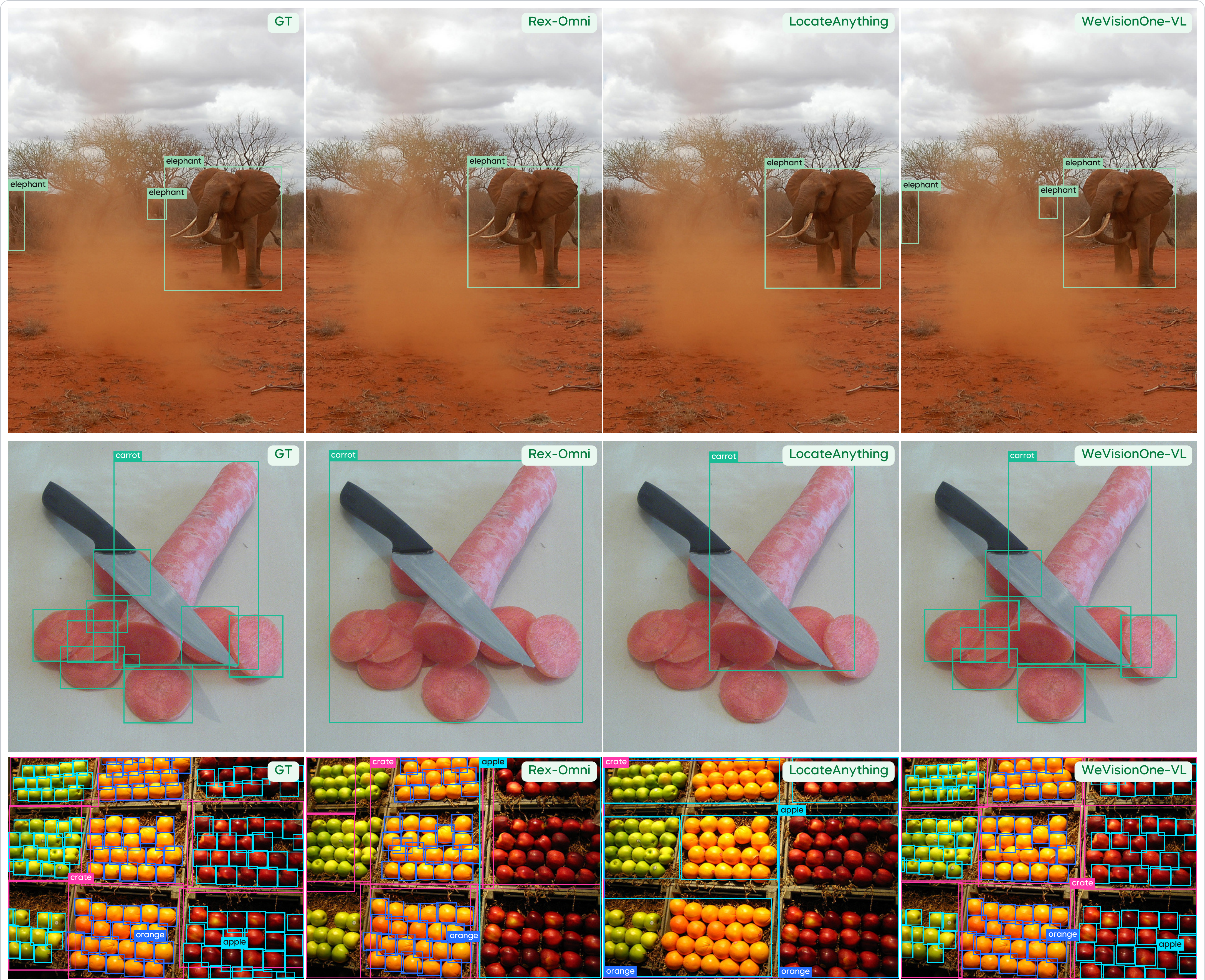}
    \end{minipage}
  \end{adjustbox}
  \par\nointerlineskip\vspace{10pt}
  \usebox{\qualanalysis}
  \par\vspace*{12pt}
\end{minipage}

\par
\Needspace*{0.9\textheight}
\noindent\begin{minipage}{\linewidth}  
  \begin{lrbox}{\qualanalysis}
    \begin{minipage}{\linewidth}
  \captionof{figure}{Qualitative comparison on referring object detection.}
  \label{fig:qual_referring}
  \vspace{9pt}
  
\textbf{Referring object detection.}
Figure~\ref{fig:qual_referring} highlights five aspects of referring object detection: spatial position, world knowledge, fine-grained attributes, negative attributes, and absent-target queries. The first four examples require interpreting ordinal position, grounding a person's name, distinguishing headphone color, and excluding people wearing sunglasses, respectively. In these cases, \modelname{} successfully selects the correct target, while the baselines select distractors or include additional people. The final query specifies a person wearing a white shirt and red shorts, but no one in the image satisfies both conditions. Both baselines still predict a box, whereas \modelname{} correctly rejects the query.
    \end{minipage}
  \end{lrbox}
  \setlength{\qualimageheight}{\dimexpr\textheight-\pagetotal-\ht\qualanalysis-\dp\qualanalysis-36pt\relax}
  \vspace*{12pt}
  \centering
  \begin{adjustbox}{max totalheight=\qualimageheight,max width=\linewidth}
    \begin{minipage}{\linewidth}
      \setlength{\parskip}{0pt}
      \includegraphics[width=\linewidth]{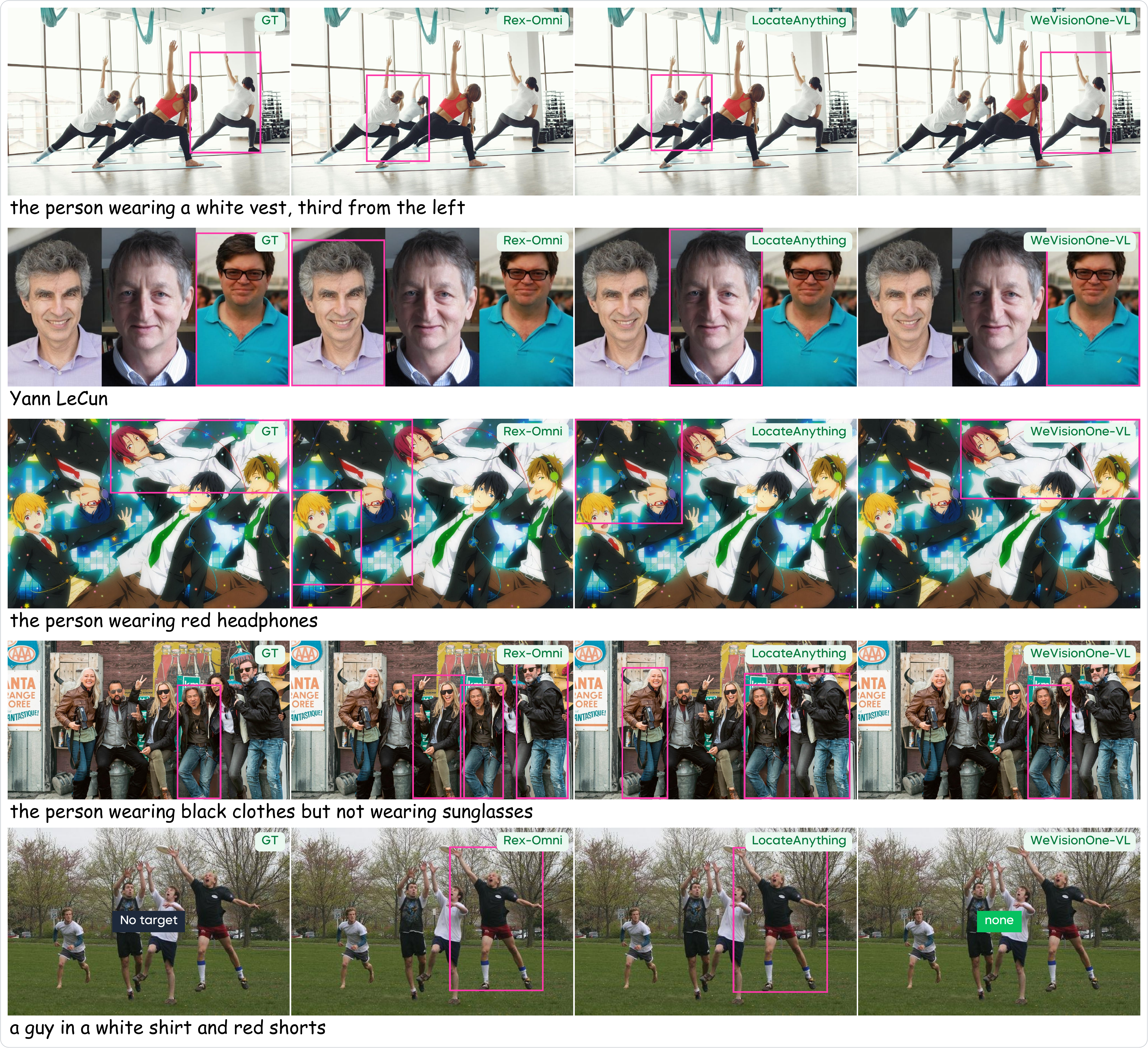}
    \end{minipage}
  \end{adjustbox}
  \par\nointerlineskip\vspace{10pt}
  \usebox{\qualanalysis}
  \par\vspace*{12pt}
\end{minipage}

\par
\Needspace*{0.9\textheight}
\noindent\begin{minipage}{\linewidth}  
  \begin{lrbox}{\qualanalysis}
    \begin{minipage}{\linewidth}
  \captionof{figure}{Qualitative comparison on object counting.}
  \label{fig:qual_counting}
  \vspace{9pt}
  
\textbf{Object counting.}
Figure~\ref{fig:qual_counting} shows the importance of the queried counting unit. The first query asks for cups on the left, and the second asks for pairs of footwear rather than individual shoes. The remaining queries specify green balls and fake green apples among distractors. The baselines often count objects outside the requested subset or count individual shoes instead of pairs; in the last example, Rex-Omni reports 537 rather than 42. \modelname{} gives the ground-truth count in all four examples.
    \end{minipage}
  \end{lrbox}
  \setlength{\qualimageheight}{\dimexpr\textheight-\pagetotal-\ht\qualanalysis-\dp\qualanalysis-36pt\relax}
  \vspace*{12pt}
  \centering
  \begin{adjustbox}{max totalheight=\qualimageheight,max width=\linewidth}
    \begin{minipage}{\linewidth}
      \setlength{\parskip}{0pt}
      \includegraphics[width=\linewidth]{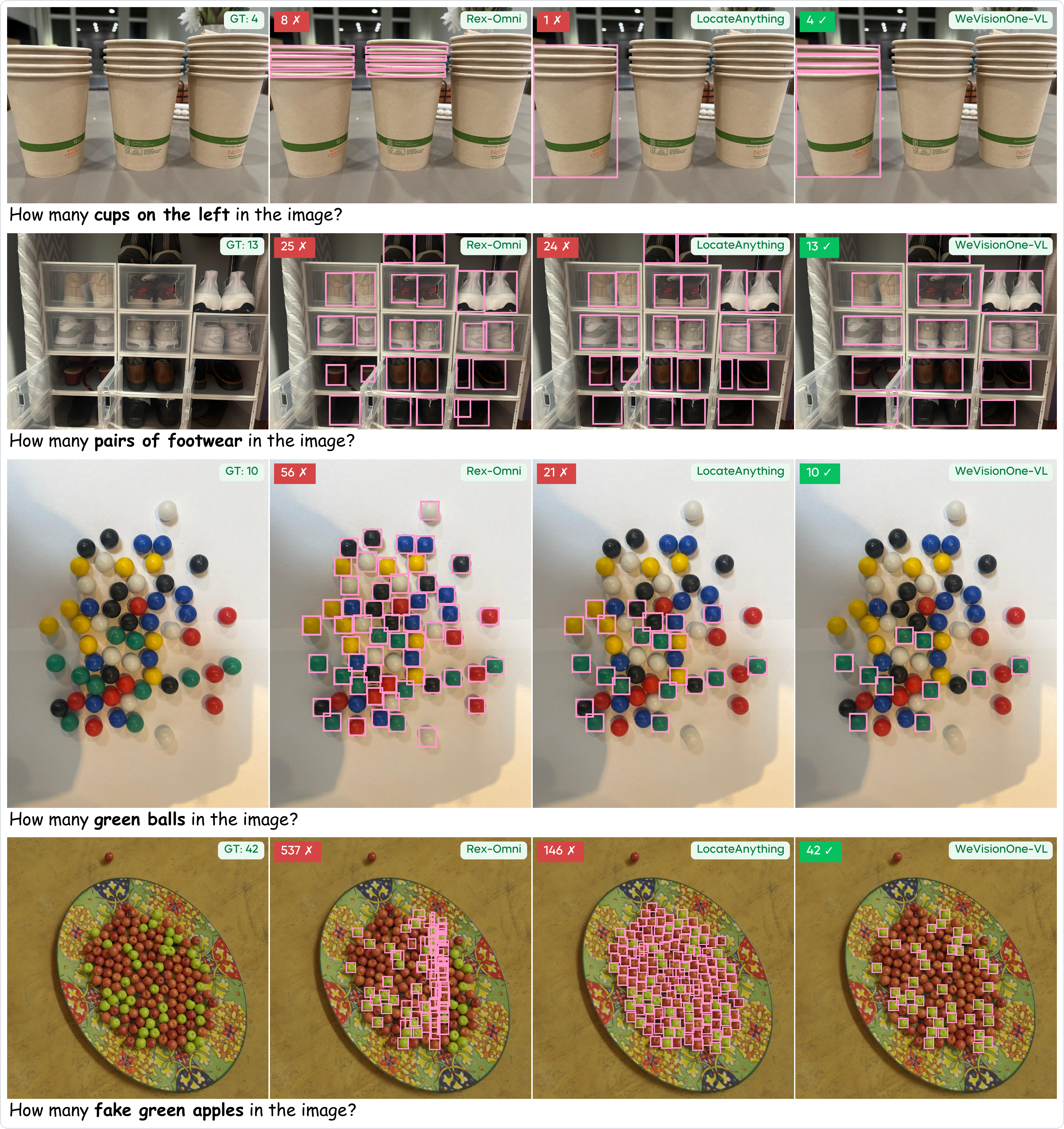}
    \end{minipage}
  \end{adjustbox}
  \par\nointerlineskip\vspace{10pt}
  \usebox{\qualanalysis}
  \par\vspace*{12pt}
\end{minipage}

\par
\Needspace*{0.9\textheight}
\noindent\begin{minipage}{\linewidth}  
  \begin{lrbox}{\qualanalysis}
    \begin{minipage}{\linewidth}
  \captionof{figure}{Qualitative comparison on object pointing. Ground-truth target regions are highlighted in green; predicted points are shown in pink.}
  \label{fig:qual_pointing}
  \vspace{9pt}
  
\textbf{Object pointing.}
Figure~\ref{fig:qual_pointing} shows dense, relational, attribute-based, and fine-grained pointing queries. In the first example, \modelname{} places points across the heart-shaped balloons, while LocateAnything misses most instances and Rex-Omni produces redundant points on some balloons. The second query asks for two people gazing into each other's eyes: both baselines select other people, whereas \modelname{} points to the indicated pair. In the third example, Rex-Omni points to multiple pigs and LocateAnything selects a distractor, while \modelname{} identifies the pig facing upward. In the final example, both baselines point to the handle, whereas \modelname{} places its point on the queried curved hook.
    \end{minipage}
  \end{lrbox}
  \setlength{\qualimageheight}{\dimexpr\textheight-\pagetotal-\ht\qualanalysis-\dp\qualanalysis-36pt\relax}
  \vspace*{12pt}
  \centering
  \begin{adjustbox}{max totalheight=\qualimageheight,max width=\linewidth}
    \begin{minipage}{\linewidth}
      \setlength{\parskip}{0pt}
      \includegraphics[width=\linewidth]{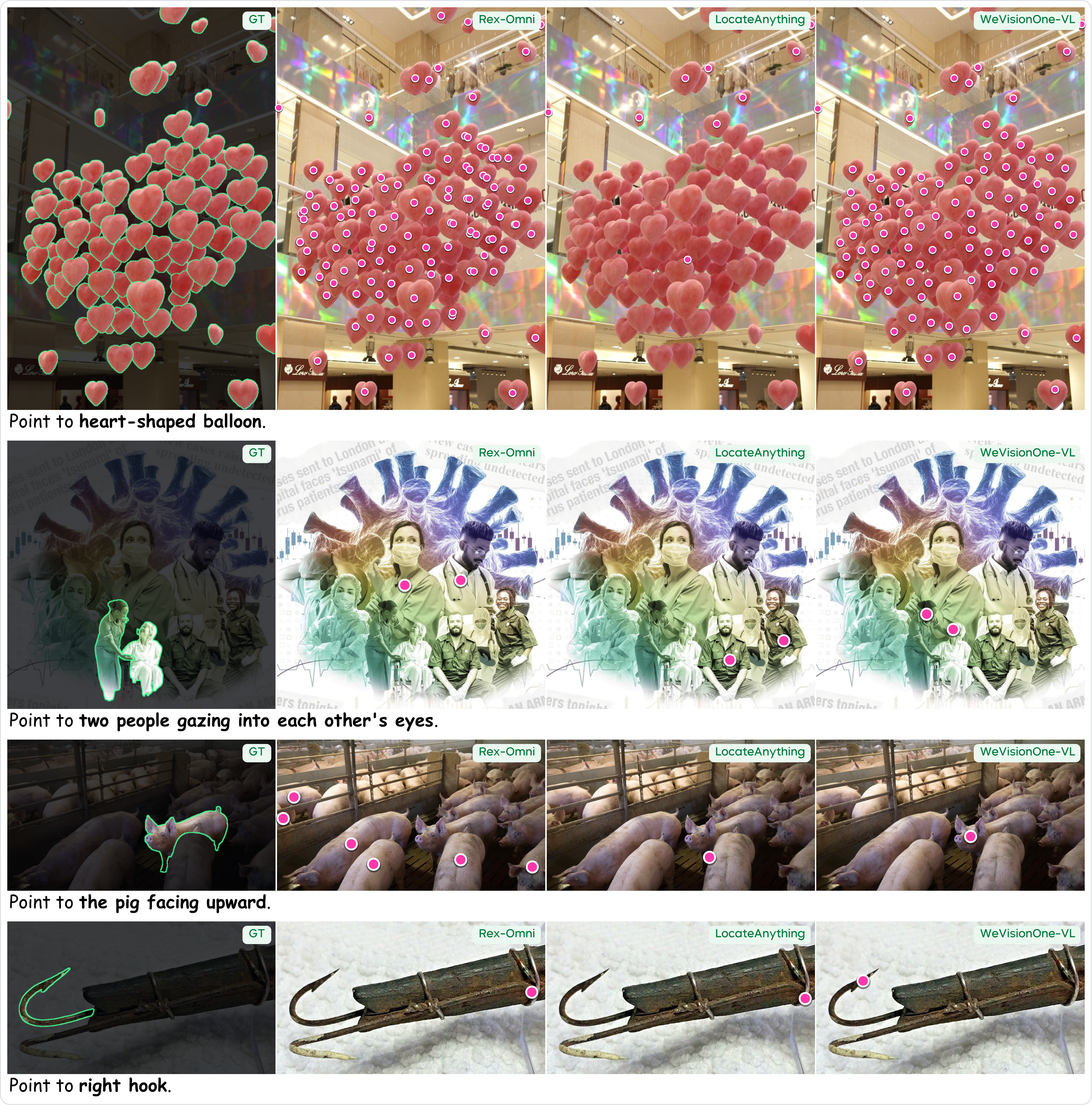}
    \end{minipage}
  \end{adjustbox}
  \par\nointerlineskip\vspace{10pt}
  \usebox{\qualanalysis}
  \par\vspace*{12pt}
\end{minipage}

\par
\Needspace*{0.9\textheight}
\noindent\begin{minipage}{\linewidth}  
  \begin{lrbox}{\qualanalysis}
    \begin{minipage}{\linewidth}
  \captionof{figure}{Qualitative comparison on OCR.}
  \label{fig:qual_ocr}
  \vspace{9pt}
  
\textbf{OCR.}
Figure~\ref{fig:qual_ocr} presents a shop sign with curved lettering, followed by a receipt and an instruction sheet with dense small text. On the shop sign, the baselines recognize some prominent words but miss the curved lettering, whereas \modelname{} localizes and transcribes the curved words. On the receipt, all methods read several lines, but \modelname{} covers more of the lower text. The difference is more pronounced on the instruction sheet, where \modelname{} localizes substantially more of the small text than either baseline.
    \end{minipage}
  \end{lrbox}
  \setlength{\qualimageheight}{\dimexpr\textheight-\pagetotal-\ht\qualanalysis-\dp\qualanalysis-36pt\relax}
  \vspace*{12pt}
  \centering
  \begin{adjustbox}{max totalheight=\qualimageheight,max width=\linewidth}
    \begin{minipage}{\linewidth}
      \setlength{\parskip}{0pt}
      \qualocrstrip{0 1942 0 0}
      \qualocrstrip{0 0 0 1391}
    \end{minipage}
  \end{adjustbox}
  \par\nointerlineskip\vspace{10pt}
  \usebox{\qualanalysis}
  \par\vspace*{12pt}
\end{minipage}

\par
\Needspace*{0.9\textheight}
\noindent\begin{minipage}{\linewidth}  
  \begin{lrbox}{\qualanalysis}
    \begin{minipage}{\linewidth}
  
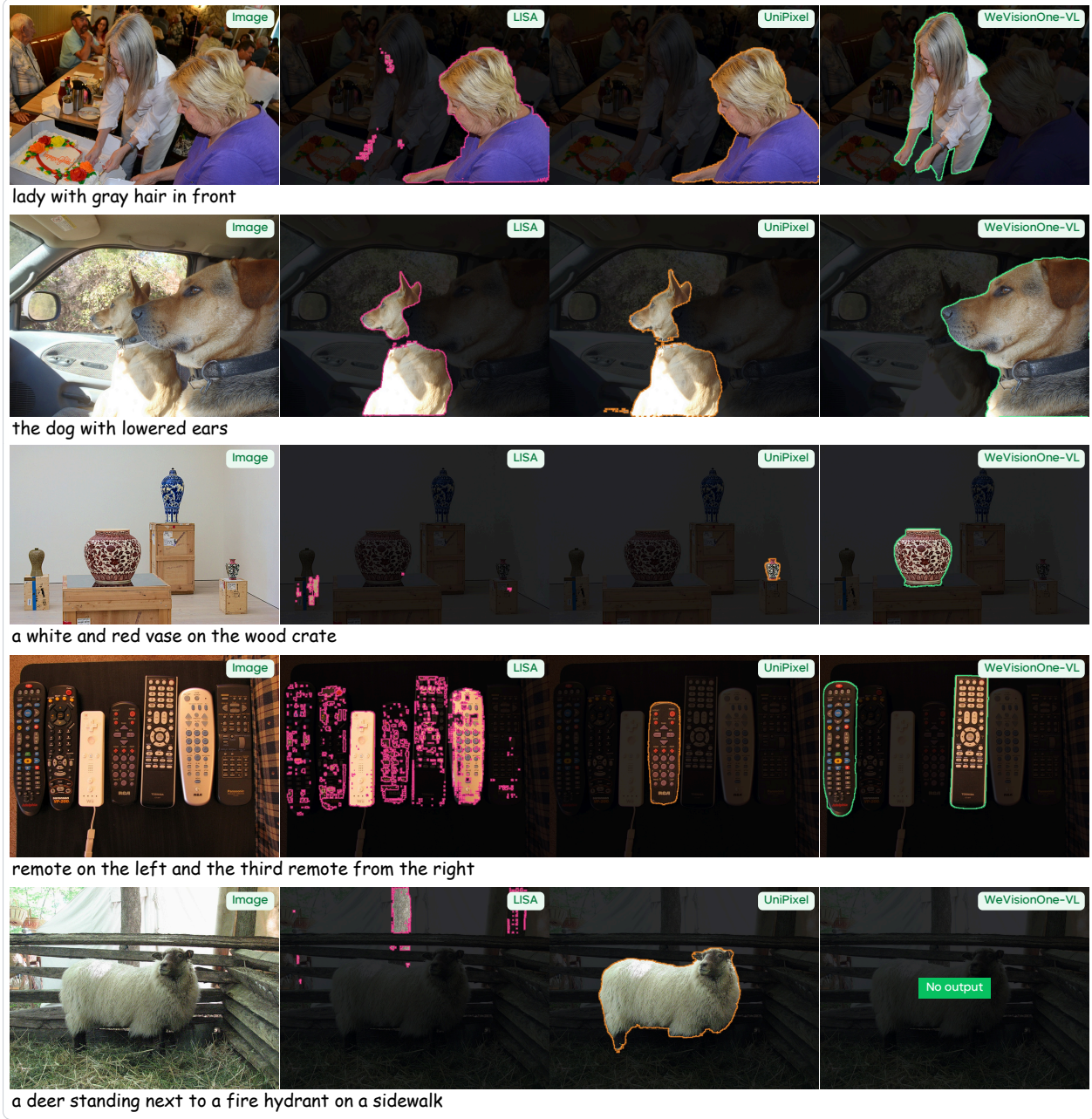
\captionof{figure}{Qualitative comparison on referring segmentation. Predicted masks remain in full color and the rest of the image is dimmed. The final query asks for an object that is not present.}
  \label{fig:qual_refseg}
  \vspace{9pt}
  
\textbf{Referring segmentation.}
Figure~\ref{fig:qual_refseg} compares the models on attribute, relational, multi-instance, and absent-target queries. In the first three examples, the expressions refer to the woman with gray hair in front, the dog with lowered ears, and the white-and-red vase on the wood crate. LISA and UniPixel select distractors or incomplete regions, whereas \modelname{} segments the referred objects. The fourth query specifies the leftmost remote and the third remote from the right; \modelname{} segments both, while LISA produces fragmented masks and UniPixel selects a different remote. The final query describes a deer beside a fire hydrant, neither of which appears in the image. Both baselines produce masks, while \modelname{} returns no mask.
    \end{minipage}
  \end{lrbox}
  \setlength{\qualimageheight}{\dimexpr\textheight-\pagetotal-\ht\qualanalysis-\dp\qualanalysis-36pt\relax}
  \vspace*{12pt}
  \centering
  \begin{adjustbox}{max totalheight=\qualimageheight,max width=\linewidth}
    \begin{minipage}{\linewidth}
      \setlength{\parskip}{0pt}
      \includegraphics[width=\linewidth]{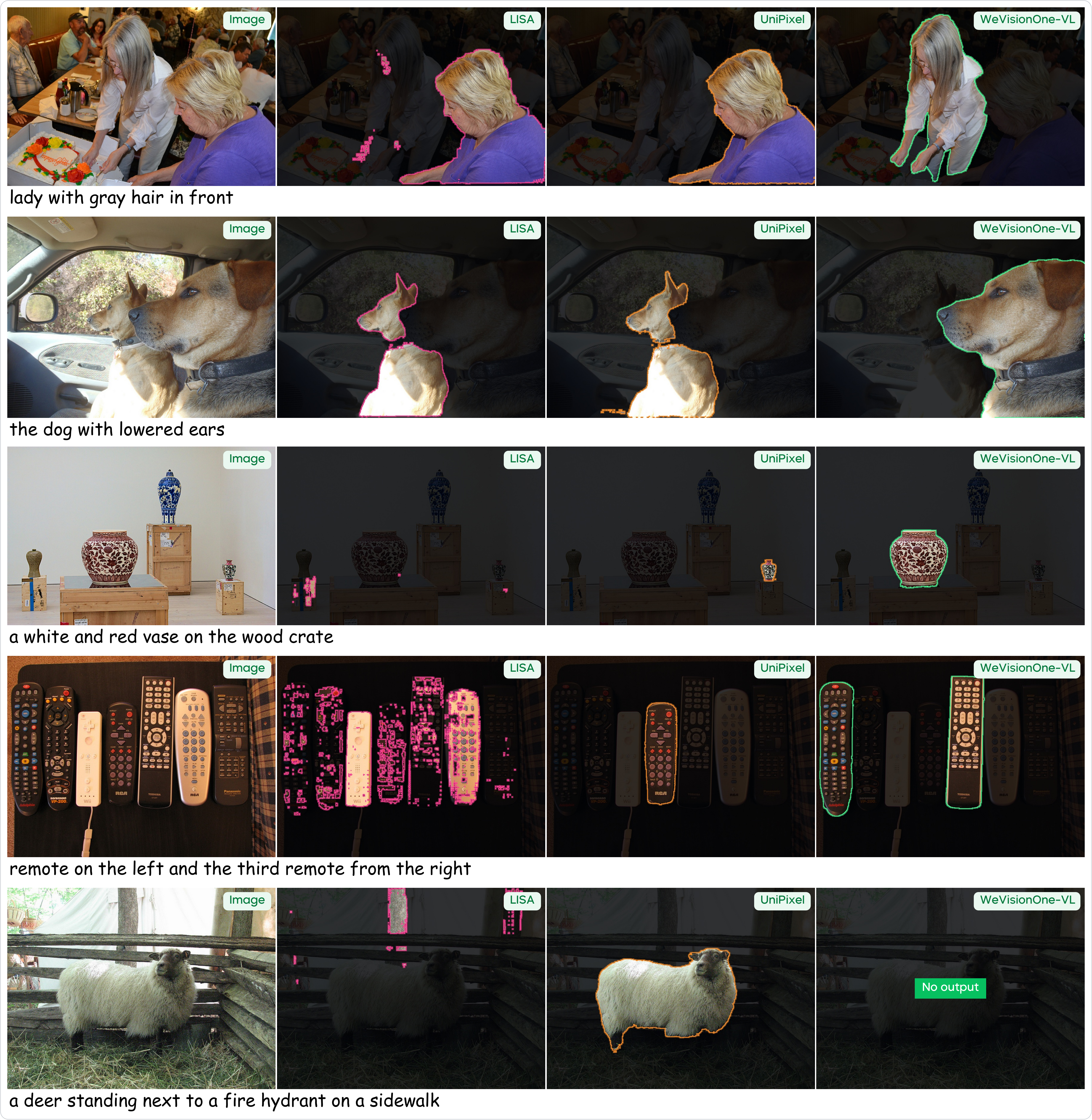}
    \end{minipage}
  \end{adjustbox}
  \par\nointerlineskip\vspace{10pt}
  \usebox{\qualanalysis}
  \par\vspace*{12pt}
\end{minipage}

\par
\Needspace*{0.9\textheight}
\noindent\begin{minipage}{\linewidth}  
  \begin{lrbox}{\qualanalysis}
    \begin{minipage}{\linewidth}
  \captionof{figure}{Qualitative comparison on box-prompted image matting.}
  \label{fig:qual_matting_box}
  \vspace{9pt}
  
\textbf{Box-prompted matting.}
Figure~\ref{fig:qual_matting_box} shows five box-prompted matting examples: a rabbit, a camel, and three people. In the first two rows, \modelname{} retains the rabbit's whiskers and more of the camel's fine coat along the boundary. The third and fifth rows highlight flyaway and curly hair strands, which appear smoother in the baseline mattes. In the fourth row, SAM2-Matting makes the apron logo transparent, leaving a hole in the subject, while \modelname{} keeps the logo and apron opaque.
    \end{minipage}
  \end{lrbox}
  \setlength{\qualimageheight}{\dimexpr\textheight-\pagetotal-\ht\qualanalysis-\dp\qualanalysis-36pt\relax}
  \vspace*{12pt}
  \centering
  \begin{adjustbox}{max totalheight=\qualimageheight,max width=\linewidth}
    \begin{minipage}{\linewidth}
      \setlength{\parskip}{0pt}
      \includegraphics[width=\linewidth]{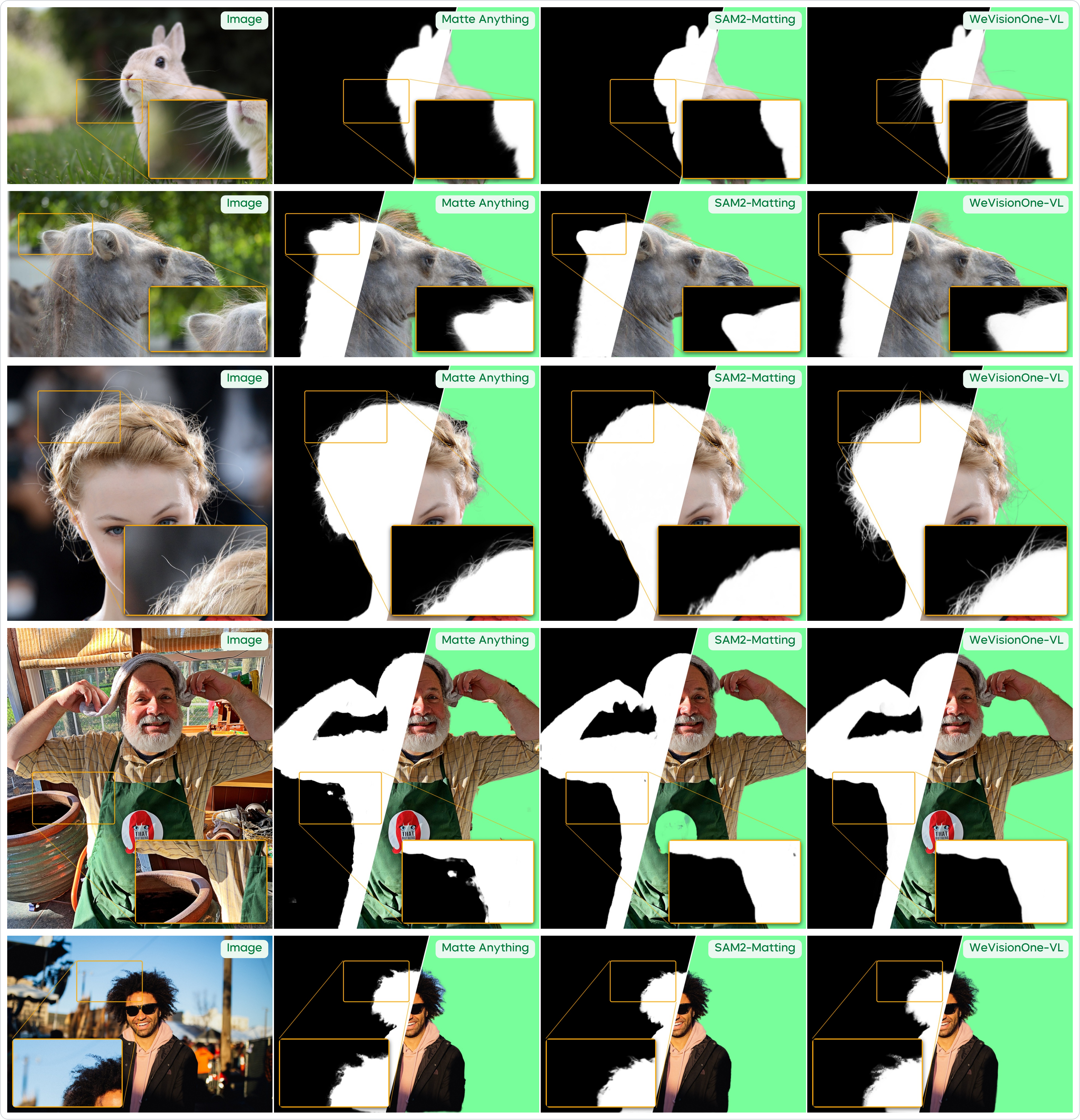}
    \end{minipage}
  \end{adjustbox}
  \par\nointerlineskip\vspace{10pt}
  \usebox{\qualanalysis}
  \par\vspace*{12pt}
\end{minipage}

\par
\Needspace*{0.9\textheight}
\noindent\begin{minipage}{\linewidth}  
  \begin{lrbox}{\qualanalysis}
    \begin{minipage}{\linewidth}
  \captionof{figure}{Qualitative comparison on text-prompted image matting.}
  \label{fig:qual_matting_text}
  \vspace{9pt}
  
\textbf{Text-prompted matting.}
Figure~\ref{fig:qual_matting_text} presents category prompts for chickens and icebergs (first two examples), followed by referring prompts for the people in black and the balloon bearing ``Daikin'' (last two). \modelname{} retains the visible chickens, gives the icebergs a more opaque matte, and mattes both intended people and the named balloon. The Qwen3.5 + SAM2-Matting pipeline selects only one of the two people in black, while SAM2-Matting (SAM3) misses them and includes several distractors in its balloon result.
    \end{minipage}
  \end{lrbox}
  \setlength{\qualimageheight}{\dimexpr\textheight-\pagetotal-\ht\qualanalysis-\dp\qualanalysis-36pt\relax}
  \vspace*{12pt}
  \centering
  \begin{adjustbox}{max totalheight=\qualimageheight,max width=\linewidth}
    \begin{minipage}{\linewidth}
      \setlength{\parskip}{0pt}
      \includegraphics[width=\linewidth]{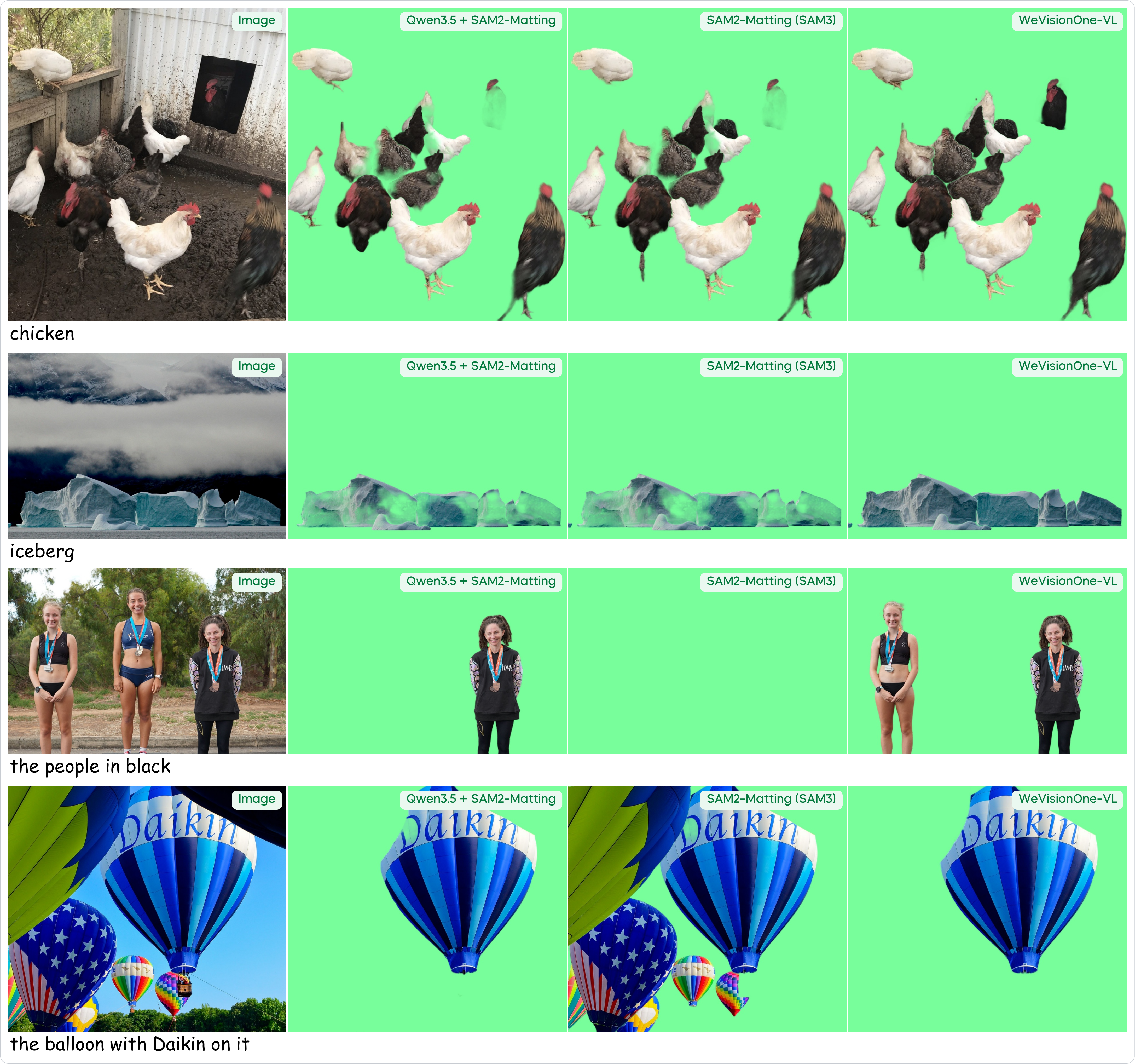}
    \end{minipage}
  \end{adjustbox}
  \par\nointerlineskip\vspace{10pt}
  \usebox{\qualanalysis}
  \par\vspace*{12pt}
\end{minipage}

\par
\Needspace*{0.9\textheight}
\noindent\begin{minipage}{\linewidth}
  \begin{lrbox}{\qualanalysis}
    \begin{minipage}{\linewidth}
  \captionof{figure}{Fine-grained caption generation with bounding boxes. Colored phrases and coordinate tokens in each caption correspond to bounding boxes of the same color in the image.}
  \label{fig:qual_grounded_caption}
  \vspace{9pt}
\textbf{Grounded caption.}
In Figure~\ref{fig:qual_grounded_caption}, we show how \modelname{} can describe an image in detail while grounding the descriptions with bounding boxes. For example, in the baseball scene, it describes the batter's clothing and swinging action, as well as the catcher and umpire behind him, and links these descriptions to the corresponding image regions. In the outdoor-table scene, it identifies the two people and individual items on the table. Such captions could be valuable for VLM training, as they provide not only detailed descriptions but also explicit connections between text and image regions, helping models learn to associate words and phrases with specific visual content.
    \end{minipage}
  \end{lrbox}
  \setlength{\qualimageheight}{\dimexpr\textheight-\pagetotal-\ht\qualanalysis-\dp\qualanalysis-36pt\relax}
  \vspace*{12pt}
  \centering
  \begin{adjustbox}{max totalheight=\qualimageheight,max width=\linewidth}
    \begin{minipage}{\linewidth}
      \setlength{\parskip}{0pt}
      \includegraphics[width=\linewidth]{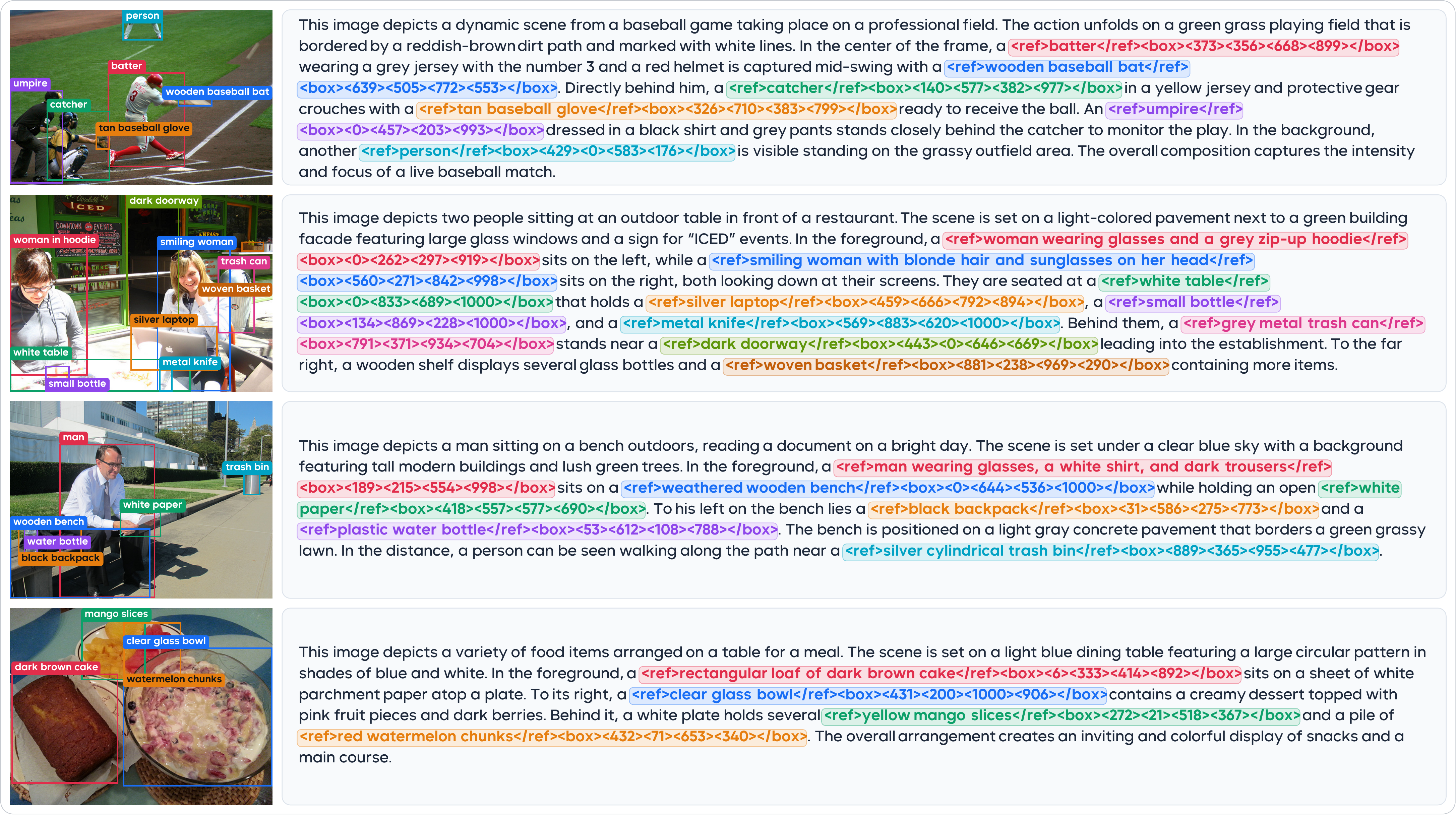}
    \end{minipage}
  \end{adjustbox}
  \par\nointerlineskip\vspace{10pt}
  \usebox{\qualanalysis}
  \par\vspace*{12pt}
\end{minipage}

\endgroup

%% file: sections/07_related_work.tex
\section{Related Work}
\label{sec:related-work}

\textbf{Visual Grounding.}
Open-vocabulary detectors~\cite{liu2024grounding,guan2025text,liang2025referdino} align object localization with category names and referring expressions. Grounded multimodal models then brought spatial outputs into a language interface. Kosmos-2~\cite{peng2023kosmos2} represents text--box links with location tokens, while Florence-2~\cite{xiao2024florence2} casts detection, grounding, and other vision tasks as prompt-conditioned sequence generation. Molmo/PixMo~\cite{deitke2024molmo} further showed the value of explicit point supervision for grounding and counting. More recently, Rex-Omni~\cite{jiang2026detect} uses quantized coordinate tokens and geometry-aware reinforcement learning for detection, referring, pointing, and OCR. LocateAnything~\cite{wang2026locateanything} decodes whole boxes and points in parallel to improve grounding speed and precision. 

\textbf{Referring Segmentation.}
SAM~\cite{kirillov2023segment} and SAM 2~\cite{ravi2025sam2} established promptable mask prediction, and SAM 3~\cite{carion2026sam3} extends this line to concept-prompted detection, segmentation, and tracking. To connect language reasoning with pixels, LISA~\cite{lai2024lisa} maps a generated segmentation token to a mask decoder, GLaMM~\cite{rasheed2024glamm} interleaves text and grounded masks, and GSVA~\cite{xia2024gsva} supports multiple referents and explicit rejection of absent targets. Later work explores different output representations and task mixtures: Text4Seg~\cite{lan2024text4seg} serializes masks as text, and UniPixel~\cite{liu2026unipixel} integrates masks and pixel-level reasoning. Sa2VA~\cite{yuan2026sa2va} couples an MLLM with SAM 2, while Youtu-VL~\cite{wei2026youtu} trains visual and language tokens under unified autoregressive supervision. 

\textbf{Instance Matting.}
Image matting estimates continuous opacity for fine structures and translucent boundaries. Automatic methods such as MODNet~\cite{ke2022modnet} and P3M~\cite{li2021privacy} focus on portrait foregrounds. Matte Anything~\cite{yao2024matte} and Matting Anything~\cite{li2024matting} use promptable segmentation to guide alpha prediction, while In-Context Matting~\cite{guo2024incontextmatting} specifies the foreground through a prompted reference image. SAM2-Matting~\cite{shen2026sam2matting} adds matting components to a segmentation tracker. For video, MatAnyone~\cite{yang2025matanyone} propagates the matte of a target specified by a first-frame mask. MatAnyone 2~\cite{yang2026matanyone2} improves detail and temporal consistency through learned quality evaluation and larger real-world training data. Many matting evaluations assume a salient foreground~\cite{ke2022modnet,li2021privacy} or provide a ground-truth mask prompt~\cite{yang2025matanyone}. We evaluate promptable matting in multi-object scenes using boxes, category names, and referring expressions to specify targets, with \modelname{} predicting their continuous alpha mattes.

%% file: sections/08_conclusion.tex
\section{Limitations}
In this work, we focus on fundamental visual perception, which entails several trade-offs. First, the limited amount of general-purpose training data may compromise the general capabilities relative to the base model. Second, for inference efficiency, we train the model to directly produce structured outputs without chain-of-thought (CoT) reasoning~\cite{liu2025segzero,feng2025onethinker}, which potentially limits its ability to tackle complex reasoning tasks. Third, our current approach is limited to single-turn prediction. It is promising to integrate an agent framework to progressively refine visual understanding through multi-turn tool interactions~\cite{jiang2026referagent,liang2026segresearch}. 

Beyond these limitations, a key challenge lies in the real-world complexity. The object prompts in benchmarks are typically clear, well-formed, and unambiguous. However, real-world user instructions are often vague and hesitant. Variations in contexts or communication styles can make a straightforward task, such as locating the target, significantly more challenging. Therefore, strong intent reasoning is essential for a reliable vision system. In future work, we will continue to improve \modelname toward addressing these limitations.

\section{Conclusion}
\label{sec:conclusion}

We introduced \modelname{}, a vision-language model that unifies object-level (e.g., object detection, referring localization, counting, and OCR) and pixel-level (e.g., referring segmentation, and image matting) tasks through a structured next-token interface. Dedicated coordinate tokens encode instance predictions, while an \texttt{<OBJ>} token connects each localized instance to a lightweight pixel decoder for masks and alpha mattes. A three-stage supervised curriculum develops task alignment, high-resolution dense perception, and pixel-level prediction. After that, we apply mixed-task reinforcement learning to optimize the complete grounding outputs. Across the evaluated benchmarks, \modelname{} achieves strong results on both object-level and pixel-level tasks. It improves mean detection F1 and counting accuracy over comparable-scale VLMs, advances text grounding and referring localization, and surpasses the evaluated baselines for referring segmentation and image matting. Qualitative examples further show that the model can localize individual instances in crowded scenes, ground expressions involving multiple objects, and recognize when a requested target is absent. By unifying these capabilities within a single language-guided model, \modelname{} provides a foundation for practical visual systems.

%% file: sections/09_appendix.tex
\appendix

\section*{Appendix}

This appendix describes the training corpus of \modelname{} and provides implementation details.

\section{\modelname{} Training Corpus}
\label{sec:appendix_data_composition}

We describe the data used for SFT, segmentation and matting training, and RL below.

\subsection{SFT Data}
\label{sec:appendix_sft_data}

Table~\ref{tab:appendix_sft_data} summarizes our SFT data by task and annotation source, including public annotations and annotations generated by the \modelname Data Engine (WDE).

\begin{table}[H]
\centering
\caption{Training data volumes across the three SFT stages (M).}
\label{tab:appendix_sft_data}
\ExpTableSetup
\small
\setlength{\tabcolsep}{5pt}
\renewcommand{\arraystretch}{1.2}
\renewcommand{\tabularxcolumn}[1]{m{#1}}
\begin{tabularx}{\linewidth}{@{}>{\raggedright\arraybackslash}m{.13\linewidth}>{\centering\arraybackslash}m{.09\linewidth}>{\raggedright\arraybackslash}X*{3}{>{\centering\arraybackslash}m{.045\linewidth}}@{}}
\toprule
\rowcolor{OursRowColor}
Task & Source & Datasets & S1 & S2 & S3 \\
\midrule
\multirow{2}{*}{\textbf{Detection}} & Public & Objects365~\cite{shao2019objects365}, OpenImages-V6~\cite{kuznetsova2020openimages}, V3Det~\cite{wang2023v3det}, DeepFashion2~\cite{ge2019deepfashion2}, ODinW-35~\cite{li2022elevater}, EgoObjects~\cite{zhu2023egoobjects}, PACO~\cite{ramanathan2023paco}, PartImageNet++~\cite{li2024pinpp}, BDD100K~\cite{yu2020bdd100k}, nuImages~\cite{nuimages2020}, VisDrone~\cite{du2019visdrone}, CrowdHuman~\cite{shao2018crowdhuman}, SKU-110K~\cite{goldman2019precise}, PartImageNet~\cite{he2021partimagenet}, MOT20Det~\cite{dendorfer2020mot20}, MOT17Det~\cite{dendorfer2020motchallenge}, OWDOD~\cite{shubh3032025owdod}, FSCD-147~\cite{nguyen2022fewshot} & 18.4 & 16.2 & 8.4 \\
\addlinespace[2pt]
\cmidrule(lr){2-6}
\addlinespace[2pt]
 & WDE & CC3M, GRIT, LAION, SA-1B & 42.2 & 10.7 & 7.2 \\
\midrule
\multirow{2}{*}{\textbf{Referring}} & Public & RefCOCO~\cite{yu2016modeling}, gRefCOCO~\cite{he2023grec}, Flickr30k~\cite{plummer2015flickr30k}, HumanRef-CoT~\cite{jiang2026rexthinker}, HumanParts~\cite{li2019detector}, $D^3$~\cite{xie2023described}, OpenRef~\cite{wu2026openref}, Unsplash~\cite{wang2026locateanything} & 6.3 & 2.1 & 1.0 \\
\addlinespace[2pt]
\cmidrule(lr){2-6}
\addlinespace[2pt]
 & WDE & GRIT, CC3M, LAION & 11.8 & 5.5 & 1.6 \\
\midrule
\textbf{Counting} & Public & CLOC~\cite{lei2026countanything}, MixCount~\cite{dumery2026mixcount}, KubriCount~\cite{liu2026countanythinggranularity}, PixMo~\cite{deitke2024molmo}, OWDOD~\cite{shubh3032025owdod}, FSC-147~\cite{ranjan2021learning}, FSCD-147~\cite{nguyen2022fewshot} & 2.4 & 1.2 & 0.6 \\
\midrule
\textbf{Pointing} & Public & PixMo~\cite{deitke2024molmo}, Molmo2~\cite{clark2026molmo2}, CLOC~\cite{lei2026countanything}, FSC-147~\cite{ranjan2021learning} & 2.8 & 0.9 & 0.4 \\
\midrule
\textbf{OCR} & Public & IDL-OCR~\cite{biten2022ocridl}, BLIP3-OCR~\cite{xue2024blip3}, ArT~\cite{chng2019art}, COCO-Text~\cite{veit2016cocotext}, LSVT~\cite{sun2019lsvt}, TextOCR~\cite{singh2021textocr}, ReCTS~\cite{zhang2019rects}, HierText~\cite{long2022towards}, RCTW~\cite{shi2017rctw}, ICDAR2015~\cite{karatzas2015icdar}, WildReceipt~\cite{sun2021wildreceipt}, SROIE~\cite{huang2019icdar2019}, ICDAR2013~\cite{karatzas2013icdar} & 2.1 & 0.7 & 0.3 \\
\midrule
\textbf{Caption} & Public & RefSpatial~\cite{zhou2025roborefer}, Inst-IT~\cite{peng2024instit}, VG-PhraseCut~\cite{wu2020phrasecut}, Flickr30k~\cite{young2014flickr30k}, COCONut-PanCap~\cite{deng2025coconutpancap}, PixMo~\cite{deitke2024molmo}, LLM-Seg40K~\cite{wang2024llmseg}, PanCap, Rea2Seg~\cite{gao2026rea2seg} & 2.7 & 0.9 & 0.4 \\
\midrule
\multirow{2}{*}[-0.5\baselineskip]{\textbf{Segmentation}} & Public & RefCOCO~\cite{yu2016modeling}, RefCOCO+~\cite{yu2016modeling}, RefCOCOg~\cite{mao2016generation}, ReVOS~\cite{yan2024visa}, Ref-YouTube-VOS~\cite{seo2020urvos} & 0.9 & 0.9 & 1.6 \\
\addlinespace[2pt]
\cmidrule(lr){2-6}
\addlinespace[2pt]
 & WDE & CC3M, GRIT, LAION, SA-1B & 0.6 & 1.4 & 20.3 \\
\midrule
\textbf{Total} & & &  \textbf{90.2} & \textbf{40.5} & \textbf{41.8} \\
\bottomrule
\end{tabularx}
\end{table}

\subsection{Segmentation and Matting Data}
\label{sec:appendix_segmentation_matting_data}

Table~\ref{tab:appendix_segmentation_matting_data} summarizes the data sources for segmentation and matting training. Public matting datasets provide high-resolution images with high-quality alpha mattes. During matting fine-tuning, we sample public and WDE-annotated data at a ratio of 1:4 within each batch.

\begin{table}[H]
\centering
\caption{Data sources and sample counts for segmentation and matting (M).}
\label{tab:appendix_segmentation_matting_data}
\ExpTableSetup
\small
\setlength{\tabcolsep}{5pt}
\renewcommand{\arraystretch}{1.2}
\renewcommand{\tabularxcolumn}[1]{m{#1}}
\begin{tabularx}{\linewidth}{@{}>{\raggedright\arraybackslash}m{.13\linewidth}>{\centering\arraybackslash}m{.09\linewidth}>{\raggedright\arraybackslash}X>{\centering\arraybackslash}m{.09\linewidth}@{}}
\toprule
\rowcolor{OursRowColor}
Task & Source & Datasets & Size (M) \\
\midrule
\multirow{2}{*}{\textbf{Segmentation}} & Public & SA-1B~\cite{kirillov2023segment}, RefCOCO~\cite{yu2016modeling}, RefCOCO+~\cite{yu2016modeling}, RefCOCOg~\cite{mao2016generation}, ReVOS~\cite{yan2024visa}, Ref-YouTube-VOS~\cite{seo2020urvos} & 171.4 \\
\addlinespace[2pt]
\cmidrule(lr){2-4}
\addlinespace[2pt]
 & WDE & CC3M, GRIT, LAION & 288.4 \\
\midrule
\multirow{2}{*}{\textbf{Matting}} & Public & AM-2K~\cite{li2022bridging}, Distinctions-646~\cite{qiao2020attention}, UHRIM~\cite{lin2025memory}, HHM-50K~\cite{sun2023ultrahigh}, IHIM-50K~\cite{huynh2024maggie}, RefMatte~\cite{li2023referringmatting}, AIM-500~\cite{li2021automatic}, CelebA-Hair~\cite{cpuimage2025celebahair}, Transparent-460~\cite{cai2022transmatting}, SIMD~\cite{sun2021semantic}, P3M-10K (train)~\cite{li2021privacy}, Real-Human~\cite{wang2021crgnn}, RealWorldPortrait-636~\cite{yu2021mask} & 0.3 \\
\addlinespace[2pt]
\cmidrule(lr){2-4}
\addlinespace[2pt]
 & WDE & CC3M, SA-1B & 129.8 \\
\bottomrule
\end{tabularx}
\end{table}

\subsection{Reinforcement Learning Data}
\label{sec:appendix_rl_datasets}

We apply uncertainty-aware sample selection (Section~\ref{sec:training_pipeline}) to the SFT data using the SFT checkpoint to construct the RL training set. The final RL set contains 178K examples: 44.0\% detection, 20.8\% dense detection, 17.5\% referring localization, 7.9\% OCR, 5.8\% referring segmentation, and 4.0\% pointing. Table~\ref{tab:training_data_composition} shows the complete 178K-example RL task distribution, including the additional localization and pointing examples.

\section{Implementation Details}
\label{sec:appendix_training_details}

\subsection{Training Hyperparameters}
\label{sec:appendix_hyperparams}

Table~\ref{tab:training_hyperparams} summarizes the hyperparameters that differ across the three SFT stages and RL training. SFT-S3 refers to joint token--pixel training. We use 128 GPUs for SFT and 160 GPUs for RL. All four configurations use AdamW with a weight decay of 0.01, a maximum gradient norm of 1.0, and BF16 precision. We use cosine learning-rate decay for SFT and constant or cosine schedules for RL, depending on the configuration. For SFT, we use sequence packing and specify batch sizes by the number of tokens per GPU. OCR examples in SFT-S2 and SFT-S3 use 16--256 visual tokens to accommodate long transcriptions.

\begin{table}[H]
\centering
\caption{Training hyperparameters for SFT and RL. VLM comprises the vision encoder, projector, and LLM. Dec. denotes the pixel decoder. A dash in the learning-rate rows indicates a frozen or inactive module. SFT-S3 uses LoRA for the LLM ($r=128$, $\alpha=256$, dropout $0.05$), with fully trainable token embeddings and output head. The other modules are fully trainable. For RL, the maximum prompt and response lengths are both 4,096 tokens.}
\label{tab:training_hyperparams}
\ExpTableSetup
\normalsize
\setlength{\tabcolsep}{5pt}
\renewcommand{\arraystretch}{1.2}
\setlength{\aboverulesep}{2pt}
\setlength{\belowrulesep}{2pt}
\begin{tabularx}{\linewidth}{@{}>{\raggedright\arraybackslash}p{0.28\linewidth}*{4}{C}@{}}
\toprule
\rowcolor{OursRowColor}
Hyperparameter & SFT-S1 & SFT-S2 & SFT-S3 & RL \\
\midrule
Trainable modules & LLM & VLM & VLM + Dec. & LLM \\
\midrule
Learning rate: LLM & $2\times10^{-5}$ & $5\times10^{-6}$ & $5\times10^{-6}$ & $1\times10^{-6}$ \\
Learning rate: vision encoder & -- & $5\times10^{-7}$ & $5\times10^{-6}$ & -- \\
Learning rate: projector & -- & $5\times10^{-6}$ & $5\times10^{-6}$ & -- \\
Learning rate: pixel decoder & -- & -- & $1\times10^{-5}$ & -- \\
Warmup ratio & 5\% & 5\% & 3\% & -- \\
Training steps & 13,000 & 8,995 & 9,062 & 593 \\
\midrule
Max. sequence length & 4,096 & 8,192 & 8,192 & 4,096 + 4,096 \\
Visual tokens / image & 16--512 & 64--1,960 & 64--1,960 & 64--1,960 \\
Token budget / GPU & 28,672 & 24,576 & 24,576 & 24,576 \\
Distributed training & ZeRO-2 & ZeRO-2 & ZeRO-2 & FSDP2 + vLLM \\
\bottomrule
\end{tabularx}
\end{table}

\subsection{Pixel Decoder Training}
\label{sec:appendix_pixel_decoder_training}

\textbf{Warm-up.}
After SFT-S2, we connect the VLM to a randomly initialized pixel decoder and jointly train them for 100,000 steps with segmentation supervision. The decoder learning rate is $1\times10^{-6}$ with 1,000 warmup steps. The decoder predicts masks at $512\times512$ resolution. After warm-up, we retain only the trained decoder and \texttt{<OBJ>} embedding and attach them to the original SFT-S2 checkpoint for joint training. The other VLM weights updated during warm-up are discarded.

\textbf{Joint training.}
During SFT-S3, we alternate NTP and segmentation steps at a 1:1 ratio to jointly train the SFT-S2 checkpoint with the warmed-up decoder. The VLM and decoder learning rates are $5\times10^{-6}$ and $1\times10^{-5}$, respectively, with a 3\% warmup ratio. For each packed sequence on each GPU, we randomly sample 24 \texttt{<OBJ>} tokens and pass their corresponding representations to the decoder for segmentation prediction, while retaining the full text sequence for NTP.

\textbf{Matting fine-tuning.}
After training is complete, we freeze the VLM and fine-tune the decoder for matting for 100,000 steps at an output resolution of $1024\times1024$. We use a learning rate of $5\times10^{-7}$ for the decoder and $2.5\times10^{-6}$ for the coarse-to-fine refinement branch, with 200 warmup steps. Each GPU processes up to eight \texttt{<OBJ>} tokens. The matting loss is defined in Section~\ref{sec:training_pipeline}.

The warm-up and matting phases use AdamW with cosine learning-rate decay, a weight decay of 0.05, and a maximum gradient norm of 5.0. Both phases use BF16 precision and DDP, with a token budget of 98,304 per GPU, a maximum sequence length of 8,192, and 16--1,024 visual tokens per image.

\subsection{RL Rewards}
\label{sec:appendix_rl_rewards}

For detection, dense detection, and referring localization, we use the grounding reward defined in Section~\ref{sec:training_pipeline}. For segmentation and matting, RL also uses only the box-based grounding reward; no reward is computed from the predicted masks or alpha mattes. For pointing, we compute F1 through one-to-one matching between predicted points and ground-truth masks, counting a match when the point lies inside the mask. Referring segmentation requires each predicted box to be followed by an \texttt{<OBJ>} token, while OCR uses its transcription reward. When the target is absent, the reward is 1 only if the model predicts no object.